%% file: tois2022-main.tex
\documentclass[acmsmall]{acmart}
\usepackage{amsmath}
\usepackage{color}
\usepackage{array}
\usepackage{breqn}
\usepackage{{inputenc}}
\usepackage{balance}
\usepackage{multirow,tabularx}
\usepackage{booktabs,longtable}
\usepackage{hhline}
\usepackage[flushleft]{threeparttable}
\usepackage{algorithm}
\usepackage{algorithmic}
\usepackage{epsfig}
\usepackage{graphicx}
\usepackage{epstopdf}
\usepackage{thmtools}
\usepackage{enumitem}
\usepackage{verbatim}
\usepackage{amsfonts}
\usepackage{hyperref}
\hypersetup{colorlinks=false,linkcolor=blue,urlcolor=blue,citecolor=red}
\usepackage{xcolor}
\usepackage{subcaption}
\usepackage{wrapfig}

\newcommand\norm[1]{\left\lVert#1\right\rVert}
\newcommand{\Lapl}{\mathbf{\mathop{\mathcal{L}}}}

\newcommand{\Trans}[1]{{#1}^{\top}}

\newcommand{\Mat}[1]{\textbf{#1}}

\newcommand{\Space}[1]{\mathbb{#1}}
\newcommand{\Set}[1]{\mathcal{#1}}

\newcommand{\ie}{\emph{i.e., }}
\newcommand{\eg}{\emph{e.g., }}

\newcommand{\wrt}{\emph{w.r.t. }}
\newcommand{\cf}{\emph{cf. }}

\newcommand{\za}[1]{{\color{black}{#1}}}
\newcommand{\wx}[1]{{\color{black}{#1}}}
\newcommand{\rza}[1]{{\color{black}{#1}}}
\AtBeginDocument{%
  }

\acmJournal{JACM}
\acmVolume{37}
\acmNumber{4}
\acmArticle{111}
\acmMonth{8}

\begin{document}

\title{Robust Collaborative Filtering to Popularity Distribution Shift}

\author{An Zhang$^{\dag}$}
\footnotetext[1]{An Zhang$^{\dag}$ and Wenchang Ma$^{\dag}$ contribute equally to this work.}
\affiliation{%
  \institution{National University of Singapore}
  \city{Singapore}
  \country{Singapore}
}
\email{an_zhang@nus.edu.sg}

\author{Wenchang Ma$^{\dag}$}
\affiliation{%
  \institution{National University of Singapore}
  \city{Singapore}
  \country{Singapore}
}
\email{e0724290@u.nus.edu}

\author{Jingnan Zheng}
\affiliation{%
  \institution{National University of Singapore}
  \city{Singapore}
  \country{Singapore}
}
\email{e0718957@u.nus.edu}

\author{Xiang Wang$^{*}$}
\footnotetext[2]{Xiang Wang$^{*}$ is the corresponding author, and also affiliated with Institute of Artificial Intelligence, Institute of Dataspace, Hefei Comprehensive National Science Center.
}
\affiliation{%
  \institution{University of Science and Technology of China}
  \city{Hefei}
  \country{China}
}
\email{xiangwang1223@gmail.com}

\author{Tat-Seng Chua}
\affiliation{%
  \institution{National University of Singapore}
  \city{Singapore}
  \country{Singapore}
}
\email{dcscts@nus.edu.sg}

\renewcommand{\shortauthors}{Trovato et al.}

\input{chapters/0_abstract}


\setcopyright{acmlicensed}
\acmJournal{TOIS}
\acmYear{2023} \acmVolume{1} \acmNumber{1} \acmArticle{1} \acmMonth{1} \acmPrice{}\acmDOI{10.1145/3627159}

\maketitle

\input{chapters/1_introduction}
\input{chapters/3_method}

\input{chapters/4_justification}

\input{chapters/5_experiments}

\input{chapters/2_related_work}
\input{chapters/6_conclusion}

\begin{acks}
This research is supported by the NExT Research Center, the National Natural Science Foundation of China (9227010114), and the University Synergy Innovation Program of Anhui Province (GXXT-2022-040).
\end{acks}

\bibliographystyle{ACM-Reference-Format}
\bibliography{my-tois2022-popgo}

\end{document}

%% file: chapters/0_abstract.tex
\begin{abstract}
    % TODO: 1. CF models, representation learning is prune to encode the popularity shortcut
    % In leading collaborative filtering (CF) models, representations of users and items are prone to learn popularity bias as shortcut from the training interaction data.
    In leading collaborative filtering (CF) models, representations of users and items are prone to learn popularity bias \wx{in the training data as shortcuts.}
    % TODO: 2. popularity shortcut benefits the in-distribution performance, but fails to out-of-distribution data.
    The popularity shortcut tricks are good for in-distribution (ID) performance but poorly generalized to out-of-distribution (OOD) data, \za{\ie when popularity distribution of test data shifts \wrt the training one.}
    % TODO: 3. debiasing strategies try to distinguish popularity-agnostic and popularity-only patterns to predict user preference. At its core is to quantify the shortcut degree and mitigate them from representations.
    To close the gap, 
    % To overcome the degeneration of performance when encounter popularity distribution shift, 
    debiasing strategies try to assess the shortcut degrees and mitigate them from the representations.
    % TODO: 4. however, they suffer from two limitations: (1) only use the some statistical metrics (e.g., item frequency) to assess the shortcut degree heuristically; (2) assuming the test distribution is known in advance to measure the shortcut degree.
    However, there exist two deficiencies:
    (1) when measuring the shortcut degrees, most strategies only use statistical metrics on a single aspect (\ie item frequency on item and user frequency on user aspect), failing to accommodate the compositional degree of a user-item pair;
    (2) when mitigating shortcuts, \wx{many strategies} assume that the test distribution is known in advance.
    % TODO: 5. As a result, this result in low-quality debiased representations. Worse till, it ID and OOD.
    This results in low-quality debiased representations.
    Worse still, these strategies achieve OOD generalizability with a sacrifice on ID performance.

    In this work, we present a simple yet effective debiasing strategy, \textbf{PopGo}, which quantifies and reduces the interaction-wise popularity shortcut without any assumptions on the test data.
    It first learns a shortcut model, which yields a shortcut degree of a user-item pair based on their popularity representations.
    Then, it trains the CF model by adjusting the predictions with the interaction-wise shortcut degrees.
    By taking both causal- and information-theoretical looks at PopGo, we can justify why it encourages the CF model to capture the critical popularity-agnostic features while leaving the spurious popularity-relevant patterns out. 
    We use PopGo to debias two high-performing CF models (MF \cite{MF}, LightGCN \cite{LightGCN}) on four benchmark datasets.
    On both ID and OOD test sets, PopGo achieves significant gains over the state-of-the-art debiasing strategies (\eg DICE \cite{DICE}, MACR \cite{MACR}).
    Codes and datasets are available at \url{https://github.com/anzhang314/PopGo}.

\end{abstract}

%% file: chapters/1_introduction.tex
\section{Introduction}
% TODO: 1. Background of CF & Narrow down to Popularity bias. Why we focus on the popularity bias? Popularity bias is the shortcut features to recommend items. However, shortcuts easily fail into the spurious statistical cues in the training data, which might not hold in the testing data or new environments.

\begin{figure}[t]
	\centering
	\subcaptionbox{Distribution shift \wrt item popularity.\label{fig:hist_tencent}}{
        \includegraphics[width=\linewidth]{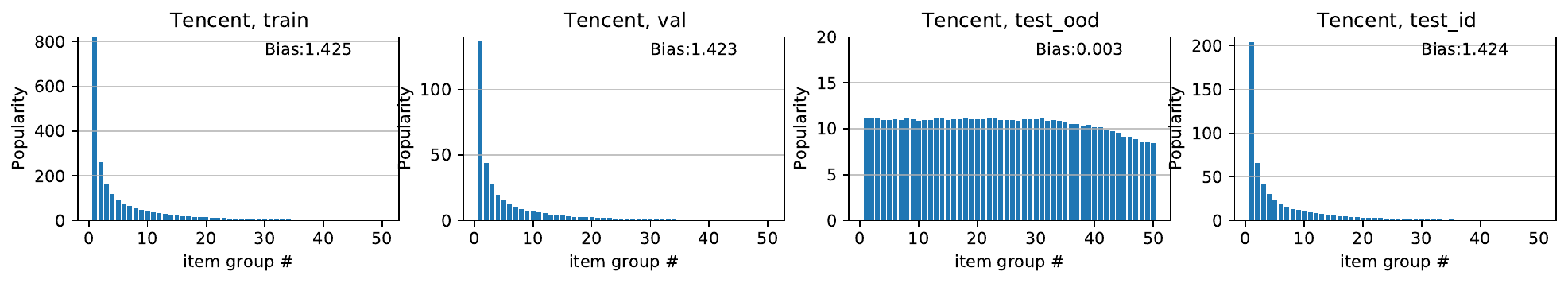}}
	\subcaptionbox{Debiasing performance on Tencent.\label{fig:intro_tencent}}{
		\includegraphics[width=0.55\linewidth]{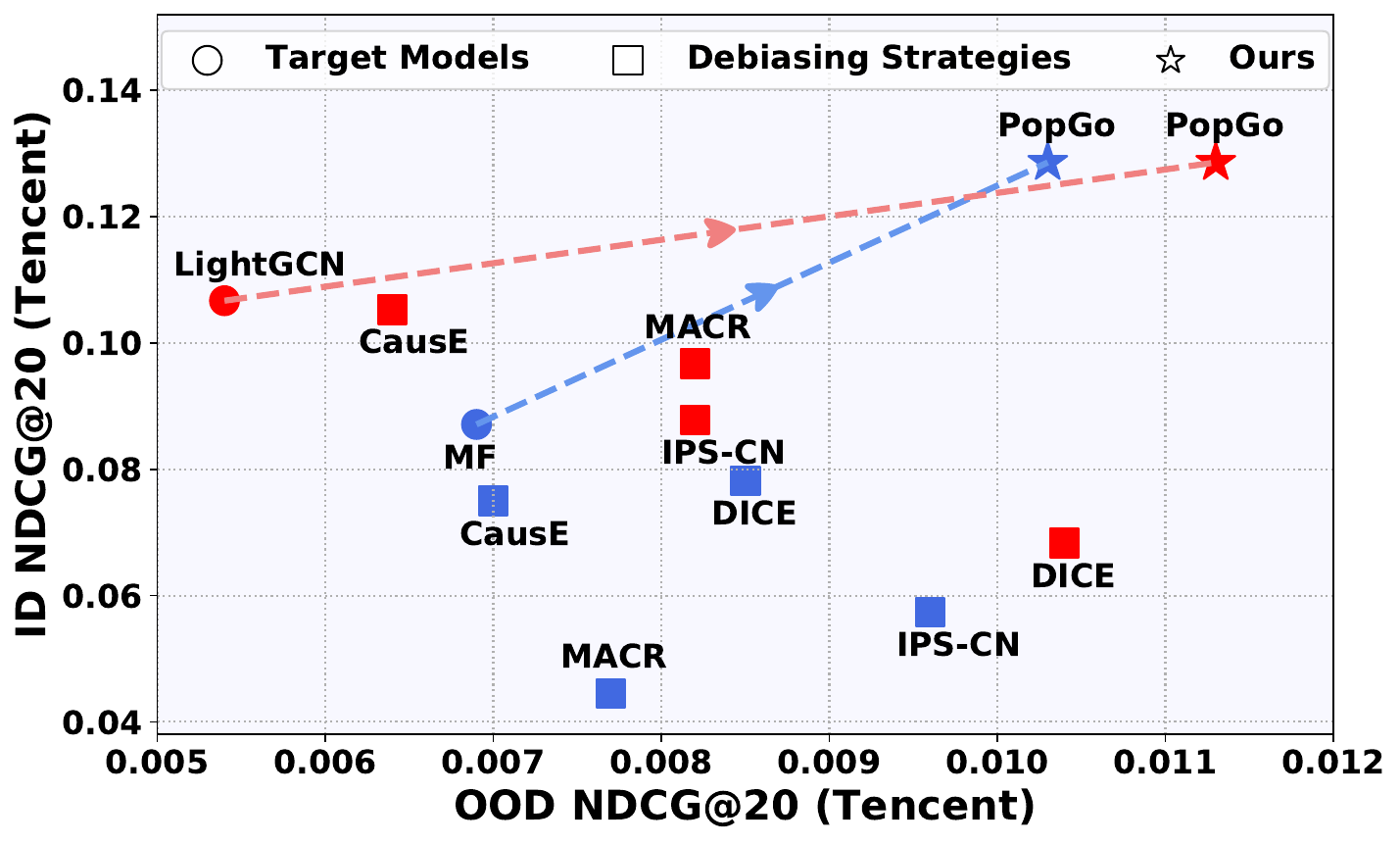}}
	\caption{(a) Popularity distribution of the training, validation, OOD testing, and ID testing sets for the Tencent dataset \cite{Tencent}. (b) Performance of debiasing strategies on the Tencent dataset, in both ID and OOD test evaluations. For strategies in blue and red, the target CF models being debiased are MF and LightGCN, respectively.}
	\label{fig:intro}
\end{figure}

Leading collaborative filtering (CF) models \cite{MF,BPR,MultVAE,LightGCN} follow a paradigm of supervised learning, which takes historical interactions among users and items as the training data, learns the representations of users and items, and consequently predicts future interactions.
Clearly, representation learning is of critical importance to delineate user preference.
% Earlier studies like matrix factorization \cite{MF,BPR} project the single identity of each user or item as an embedding.
% Some follow-on efforts, such as SVD++ \cite{SVD++} and MultVAE \cite{MultVAE}, encode the interaction history of a user or item into representations.
% Recently, one prevailing direction is organizing the training data as a holistic graph and applying graph neural networks (GNNs) \cite{Benchmarking,GraphSage,KipfW17} to incorporate multi-hop connections into representations, such as LightGCN \cite{LightGCN}.
% Despite great success, 
\wx{However,} the representations are susceptible to learning popularity bias in the training data as shortcuts.
Typically, the popularity shortcut is the superficial correlations between popularity and user preference on a particular dataset \cite{recommender_bias,Crowd,Statistical_biases,AbdollahpouriMB19,ParkT08,FPC}, but are not helpful in general.
When the test data follows the same distribution as the training data (\ie in-distribution (ID)), it is much easier to reach a high recommendation accuracy by leveraging the popularity shortcut \cite{MostPop}, instead of probing the actual preference of users.
However, the representations \wx{that} have achieved strong ID performance by utilizing shortcuts in the training dataset will generalize poorly when the shortcut is absent in the test set (\ie out-of-distribution (OOD)).
% As a result, it turns out to be a crucial problem on the generalization on CF models.
\za{Specifically, by ``OOD'' \wrt popularity shortcut, we mean that the popularity distribution shifts between the training and test data.
Considering the Tencent dataset \cite{Tencent} as an example, we split it into the training, validation, ID test, and OOD test data, and then illustrate their popularity distributions in Figure \ref{fig:hist_tencent}.
Clearly, the popularity distribution of the training data is similar to that of the ID test data, but significantly different from that of the OOD test data.
Such a divergence leads to a performance drop of CF models (\eg MF \cite{MF} and LightGCN \cite{LightGCN}), as Figure \ref{fig:intro_tencent} shows where these models perform well \wrt NDCG@20 on the ID test set, while generalizing poorly \wrt NDCG@20 on the OOD test set.}

% TODO: 2. Related Works. & 3. Limitations & Motivations.
The crucial problem on the generalization of CF models motivates the task of debiasing, which aims to alleviate the popularity bias and close the gap between ID and OOD performance.
The current in-processing debiasing strategies can be roughly categorized into five types:
(1) Inverse Propensity Scoring (IPS) \cite{IPS,UBPR,Causal_inference,IPS-CN,Propensity_SVM-Rank,AutoDebias,DR}, which views item popularity as the proxy of propensity score and exploits its inverse to reweight the loss of each user-item pair; 
% exploits the inverse of propensity score to reweight the loss of each user-item pair;
(2) Domain adaption \cite{CausE,KDCRec,ESAM}, which uses a small amount of unbiased data as the target domain to guide the model training on biased source data;
(3) Causal effect estimation \cite{DICE,MACR,DecRS,PDA,DecCaus,CauSeR,KDCRec}, which specifies the role of popularity bias in causal graphs and mitigates its effect on the prediction.
(4) Regularization-based framework \cite{ALS+Reg,Regularized_Optimization,FPC}, which explores the
use of regularization for controlling the trade-off between recommendation accuracy and coverage.
(5) Generalized methods \cite{AdvInfoNCE,bc_loss,InvCF,CD2AN,S-DRO,COR,CausPref,InvPref} learn invariant representations against popularity bias in order to achieve stability and generalization ability.

Regardless of diverse ideas, we can systematize these strategies as a standard paradigm of debiasing, which quantifies the degree of popularity shortcuts and removes them from the CF representations.
However, there exist two \wx{deficiencies}:
\begin{itemize}
    \item When measuring the shortcut degrees, most strategies \cite{IPS,Propensity_SVM-Rank,DICE,MACR} only adopt statistical metrics on a single aspect (\ie \wx{item/user frequency on item/user aspect}), and fail to accommodate the compositional degree of a user-item pair.
    For example, the IPS family \cite{IPS,Propensity_SVM-Rank} solely considers item frequency as the shortcut degree to reweight the user-item pairs, while CausE \cite{CausE} partitions user-item pairs into biased and unbiased groups based only on item frequency.
    As a result, for different users, their distinct interactions with \wx{the} same item are assigned with an identical shortcut degree, which fails to indicate the interaction-dependent shortcut.
    Hence, how to accurately identify the popularity shortcut is crucial to prevent CF models from making use of them.

    % once a popularity bias has been identified, we can improve the out-of-domain performance of models by preventing them from making use of that bias.

    \item When mitigating the shortcut, many strategies \cite{DICE,CausE,KDCRec,ESAM} predetermine that the test distribution is known in advance to guide the debiasing.
    For example, CausE \cite{CausE} involves a part of unbiased data which conforms to the test distribution during training, while MACR \cite{MACR} needs the test distribution to adjust the key hyperparameters.
    However, such assumptions are usually infeasible for real-world \wx{or wild} settings.
\end{itemize}
These limitations result in low-quality debiased representations, which hardly distinguish popularity-agnostic information from popularity shortcuts.
Worse still, as Figure \ref{fig:intro_tencent} shows, these strategies achieve higher OOD but lower ID performance, as compared with the non-debiasing CF models.
We \wx{hence argue} that the OOD improvements may result from the ID sacrifices, but not from the improved generalization ability.

In this work, we aim to truly improve the generalization by reducing popularity shortcuts and pursuing high-quality debiased representations, instead of finding a better trade-off between ID and OOD \wx{performance}.
Towards this end, we propose a simple yet effective strategy, \textbf{PopGo}, which quantifies the interaction-wise popularity shortcut without any assumptions on the test data.
\wx{Before debiasing the target CF model, it creates a shortcut model, which has the same architecture but generates shortcut representations of users and items instead.}
% Precisely, it consists of two components: (1) the target CF model, which takes the one-hot encoding of user or item identity as the input and creates the identity representation; (2) the shortcut model, which has the same architecture as the CF model, but sets the one-hot encoding of user or item popularity as the input instead and obtains the popularity representation.
We first train the shortcut model to capture the popularity-only information and yield a shortcut degree of a user-item pair based on their popularity representations.
Next, we train the CF model by adjusting its prediction on the user-item pair with \wx{its interaction-wise} shortcut degree.
As a result, the CF representations prefer to capture the critical popularity-agnostic information to predict user preference, while leaving the popularity shortcut out.
Furthermore, by taking both causal- and information-theoretical look at PopGo, we can justify why it is able to improve the generalization and remove the popularity shortcut.
We use PopGo to debias two high-performing CF models, MF \cite{BPR} and LightGCN \cite{LightGCN}, on four benchmark datasets.
\wx{In} both ID and OOD test evaluations, PopGo achieves significant gains over the state-of-the-art debiasing strategies (\eg IPS \cite{IPS}, DICE \cite{DICE}, MACR \cite{MACR}).
We summarize our main contributions as follows:
\begin{itemize}
    \item We devise a new debiasing strategy, PopGo, which quantifies the interaction-wise shortcut degrees and mitigates them towards debiased representation learning.
    \item In both causal and information theories, we validate that PopGo emphasizes the popularity-agnostic information, instead of the popularity shortcut.
    \item We conduct extensive experiments to justify the superiority of PopGo in debiasing two CF models (MF, LightGCN).
    On both ID and OOD test evaluations, PopGo significantly outperforms the state-of-the-art debiasing strategies.
\end{itemize}

%% file: chapters/3_method.tex
\section{Methodology}
\begin{figure*}[t]
    \centering
    \includegraphics[width=\linewidth]{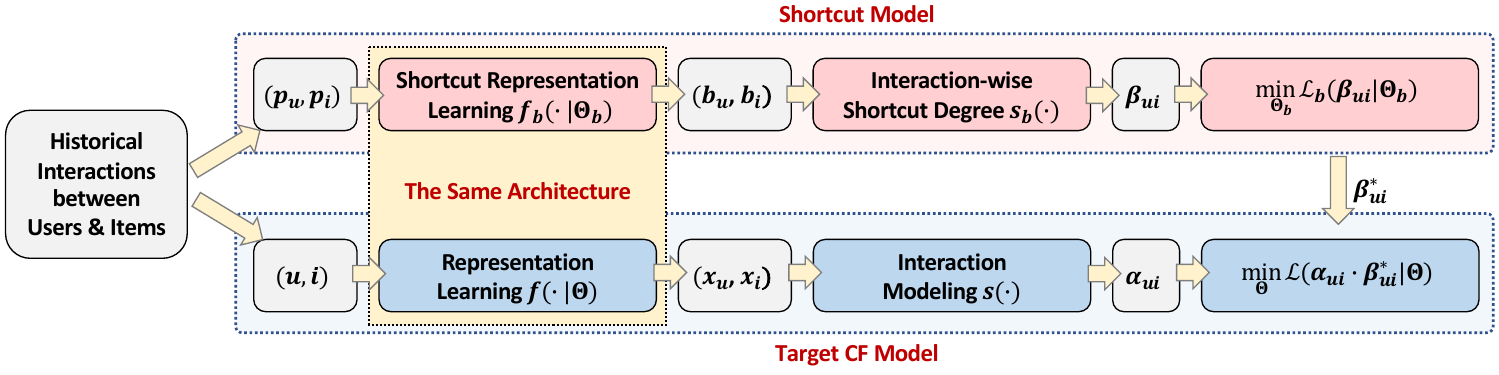}
    % \vspace{-5pt}
    \caption{The overall framework of PopGo.}
    \label{fig:framework}
    % \vspace{-5pt}
\end{figure*}

Here we first summarize the conventional learning paradigm of collaborative filtering (CF) models.
We then characterize the popularity distribution shift in CF.
We further present our debiasing strategy, PopGo, to enhance the generalization of CF models.
Figure \ref{fig:framework} illustrates the overall framework.
% On the top of the target CF model, it generates identity representations.
% With the shortcut model, it creates popularity representations to measure the shortcut degree of each user-item instance.
% Hereafter, the instance-wise shortcut adjusts the predictions upon the identity representations, so as to encourage the popularity-agnostic information.
% As a result, the target CF model approaches debiased representations.

\subsection{Conventional Learning Paradigm}
\label{sec:cf-learning-paradigm}

We focus on item recommendation from implicit feedback \cite{BPR,InfoNCE_Rendle}, where historical behaviors of a user (\eg views, purchases, clicks) reflect her preference implicitly.
Let $\Set{U}$ and $\Set{I}$ be the sets of users and items, respectively.
We denote the historical user-item interactions by $\Set{O}^{+}=\{(u,i)|y_{ui}=1\}$, where $y_{ui}$ indicates that user $u$ has adopted item $i$ before.
Under such a scenario, only identity information and historical interactions are available to deduce future interactions among users and items.

Towards this end, CF simply assumes that behaviorally similar users \wx{would} have similar preference for items.
Extensive studies \cite{BPR,MF,LightGCN} have developed the CF idea and achieved great success.
Scrutinizing these CF models, we summarize the common paradigm as a combination of two modules $z_{x}=s\circ f$: representation learning $f(\cdot)$, interaction modeling $s(\cdot)$.

\subsubsection{\textbf{Representation Learning}}
\label{sec:representation-learning}
For a user-item pair $x=(u,i)$, no overlap exists between the user identity $u$ and item identity $i$.
This semantic gap hinders the modeling of user preference.
One prevalent solution is generating informative representations of users and items, instead of superficial identities.
Here we formulate the representation learning module as a function $f(\cdot)$:
\begin{gather}\label{equ:identity-rep}
    \Mat{x}_{u},\Mat{x}_{i} = f(x=(u,i)|\Theta),
\end{gather}
where $\Mat{x}_{u}\in\Space{R}^{d}$ and $\Mat{x}_{i}\in\Space{R}^{d}$ are the $d$-dimensional representations of user $u$ and item $i$ respectively, which are expected to delineate their intrinsic characteristics;
$f(\cdot)$ is parameterized with $\Theta$.
% Parameterized with $\Theta$, $f(\cdot)$ takes the identity features $x=(u,i)$ as the input and outputs the identity representations.
As a result, this module converts $u$ and $i$ into the same latent space and closes the semantic gap.

One evolution of representation learning is in gradually bringing higher-order connections among users and items together.
Earlier studies (\eg MF \cite{MF,BPR}, NMF \cite{NCF}, CMN \cite{CMN}) project the identity of each user or item into vectorized embedding individually.
Later on, some efforts (\eg SVD++ \cite{MF}, FISM \cite{FISM}, MultVAE \cite{MultVAE}) view the historical interactions of a user (or item) as her (or its) features and integrate their embeddings into representations.
From the perspective of user-item interaction graph, interacted items of a user compose her one-hop neighbors.
Recently, some follow-on works (\eg PinSage \cite{PinSage}, LightGCN \cite{LightGCN}) apply GNNs on the user-item interaction graph holistically to incorporate multi-hop connections into representations.

\vspace{5pt}
\noindent\textbf{Discussion.}
Despite great success, evolving from modeling the single identity, the one-hop connection to the holistic interaction graph amplifies the popularity bias in the training interaction data \cite{Navip, APDA}.
Specifically, from the view of information propagation, as active users and popular items appear more frequently than the tails, they will propagate more information to the one- and multi-hop neighbors, thus contributing more to the representation learning \cite{FPC,AbdollahpouriMB19}.
In turn, the representations over-emphasize the users and items with high popularity.

\subsubsection{\textbf{Interaction Modeling}}
Having established user and item representations, we now generate the prediction on the user-item pair $x=(u,i)$.
Here we summarize the prediction function $s(\cdot)$ as:
\begin{gather}\label{equ:identity-predict}
    \alpha_{ui} = s(\Mat{x}_{u}, \Mat{x}_{i}),
\end{gather}
where $\alpha_{ui}\in\Space{R}$ reflects how likely user $u$ would adopt item $i$.
Typically, $s(\cdot)$ casts the predictive task as estimating the similarity between user representation $\Mat{x}_{u}$ and item representation $\Mat{x}_{i}$.
While neural networks like multilayer perceptron (MLP) are applicable in implementing $s(\cdot)$ \cite{NCF}, recent studies \cite{DBLP:conf/recsys/RendleKZA20,InfoNCE_Rendle} suggest that simple functions, such as inner product $\Trans{\Mat{x}}_{u}\Mat{x}_{i}$, are more suitable for real-world scenarios, due to the simplicity and efficiency.
To further convert the inner product into the probability of $u$ selecting $i$, we can resort to cosine similarity $\Trans{\Mat{x}}_{u}\Mat{x}_{i}/(\norm{\Mat{x}_{u}}\cdot\norm{\Mat{x}_{i}})$ or sigmoidal inner product $1/(1+\exp{(\Trans{\Mat{x}}_{u}\Mat{x}_{i})})$.

% We can further convert the inner product into the probability of $u$ selecting $i$ via cosine similarity $\Trans{\Mat{x}}_{u}\Mat{x}_{i}/\norm{\Mat{x}_{u}}\norm{\Mat{x}_{i}}$ or sigmoidal inner product $1/(1+\exp{(\Trans{\Mat{x}}_{u}\Mat{x}_{i})})$.

To optimize the model parameters, the current CF models mostly opt for the supervised learning objective.
Mainstream objectives can be categorized into three groups: pointwise loss (\eg binary cross-entropy \cite{NCF}), pairwise loss (\eg BPR \cite{BPR}), and softmax loss \cite{InfoNCE_Rendle}.
Here we minimize sampled softmax loss \cite{SGL} to encourage the predictive score of each observed pair, while stifling that of unobserved pairs as follows:
\begin{gather}\label{equ:identity-loss}
    \min_{\Theta}\Lapl_{0} = -\sum_{(u,i)\in\Set{O}^{+}}\log\frac{\exp{(\alpha_{ui}/\tau)}}{\sum_{j\in\Set{N}^{+}_{u}}\exp{(\alpha_{uj}/\tau)}},
\end{gather}
where $(u,i)\in\Set{O}^{+}$ is one observed interaction of user $u$, while $\Set{N}_{u}=\{j|y_{uj}=0\}$ is the sampled unobserved items that $u$ did not interact with before, and $\Set{N}^{+}_{u}=\{i\}\cup{\Set{N}_{u}}$; $\tau$ is the hyper-parameter known as the temperature in softmax.

\vspace{5pt}
\noindent\textbf{Discussion.}
Nonetheless, the learning objective is easily influenced by the popularity bias of the training interactions.
Obviously, active users and popular items dominate the historical interactions $\Set{O}^{+}$.
As a result, a lower loss can be naively achieved by recommending popular items \cite{MACR} --- the loss will optimize the model parameters towards the pairs with high popularity and inevitably influence the representations via backpropagation \cite{FPC,ZhuWC20}.

\za{
\subsection{Popularity Distribution Shift}
Here we get inspiration from the recent studies on OOD \cite{FineOOD,OoD-Bench} and take a closer look at the popularity of a user $u$, termed $d_u$.
Formally, we denote the popularity distributions by $p_{\text{train}}(d_{u})$ and $p_{\text{test}}(d_{u})$, which statistically summarize the numbers of $u$-involved interactions on the training and test sets, respectively.
Analogously, given an item $i$, we can get the popularity distributions $p_{\text{train}}(d_{i})$ and $p_{\text{test}}(d_{i})$.
As the future interactions are unseen during the training time, it is unrealistic to ensure the future (test) data distributions conform to the training data.
Thus, the popularity distribution shifts easily arises:
\begin{gather}
    p_{\text{train}}(d_{u})\neq p_{\text{test}}(d_{u}),\quad p_{\text{train}}(d_{i})\neq p_{\text{test}}(d_{i}).
\end{gather}

Worse still, as Equation \eqref{equ:identity-loss} shows that user and item representations are optimized on the training interactions $\Set{O}^{+}$, but are blind to the future (test) interactions $\Set{O}^{+}_{\text{test}}$, this learning paradigm is prone to capture the spurious correlations between the popularity and user preference.
Taking a user-item pair $(u,i)$ as an example, the spurious correlation arises when the popularity and user preference are strongly correlated at training time, but the testing phase is associated with different correlations since the popularity distribution changes.
Such a spurious correlation can be formalized as:
\begin{gather}
    p_{\text{train}}(y_{ui}|d_{u},d_{i})\neq  p_{\text{test}}(y_{ui}|d_{u},d_{i}),
\end{gather}
which easily leads to poor generalization \cite{FineOOD,OoD-Bench}.
Therefore, it is critical to make CF models robust to such challenges.
}

\subsection{PopGo Debiasing Strategy}
As discussed above, the conventional learning paradigm makes both representation learning and interaction modeling modules susceptible to learning popularity bias in the training data as the shortcut.
When the test data follows the same distribution as the training data, the popularity shortcut still holds and tricks the CF models in good ID performance.
It is much easier to reach a high recommendation accuracy by memorizing the popularity shortcut, instead of probing the actual preference of users.
However, in open-world scenarios, the wild test distribution is usually unknown and violates the training distribution, where the popularity shortcut is absent.
Hence, the CF models will generalize poorly to the OOD test data.

To enhance the generalization of CF models, it is crucial to alleviate the reliance on popularity shortcuts.
Towards this end, PopGo inspects the learning paradigm in Section \ref{sec:cf-learning-paradigm} and performs debiasing in both representation learning and interaction modeling components.
For the target CF model being debiased, PopGo devises a shortcut model, which has the same architecture as the target CF model, formally $z_{b}=s_{b}\circ f_{b}$, but functions differently:
$f_{b}(\cdot)$ focuses on shortcut representation learning, and $s_{b}(\cdot)$ targets at the modeling of interaction-wise shortcut degree.
We will elaborate on each component in the following sections.

\subsubsection{\textbf{Shortcut Representation Learning}}
\label{sec:shortcut-rep-learn}
For a user-item pair $(u,i)$, we denote user popularity and item popularity by $d_{u}$ and $d_{i}$, respectively.
Based on the statistics of the training data, $d_{u}$ is the amount of historical items that user $u$ interacted with before, while $d_{i}$ is the amount of observed interactions that item $i$ is involved in.
Although such statistical measures can be used as the influence of popularity \cite{MACR,bell2008bellkor}, they are superficial and independent of CF models, thus failing to reflect the changes after the representation learning module $f(\cdot)$ (\cf Section \ref{sec:representation-learning}).
As a result, the superficial features are insufficient to represent the popularity-relevant information.

Here we devise a function $f_{b}(\cdot)$, which has the same architecture as $f(\cdot)$ but takes the superficial features $b=(d_{u},d_{i})$ as the input.
It aims to better encapsulate the popularity-relevant information into shortcut representations:
\begin{gather}
    \Mat{b}_{u},\Mat{b}_{i} = f_{b}((d_{u},d_{i})|\Theta_{b}),
\end{gather}
where $\Mat{b}_{u}\in\Space{R}^{d}$ and $\Mat{b}_{i}\in\Space{R}^{d}$ separately denote the $d$-dimensional shortcut representations of $u$ and $i$;
$\Theta_{b}$ collects the trainable parameters of $f_{b}(\cdot)$.

Assuming MF is the target CF model that projects the one-hot encoding of the user or item identity as an embedding, a separate MF is the shortcut model that maps the one-hot encoding of the user or item frequency into a shortcut embedding.
It is worth mentioning that the frequency is treated as a categorical feature, such that the users/items with the same popularity share the identical shortcut embedding.
Analogously to the target LightGCN model, an additional LightGCN serves as the shortcut model to create the popularity embedding table and apply the GNN on it to obtain the final shortcut representations.
Note that the shortcut representations are different from the conformity embeddings of DICE \cite{DICE}, which are customized for individual users and items and hardly make the best of statistical popularity features.
Moreover, with the same target model (\eg MF, LightGCN), PopGo owns much fewer parameters than DICE.

\subsubsection{\textbf{Interaction-wise Shortcut Modeling}}
Having obtained shortcut representations of users and items, we now approach the interaction-wise short degree:
\begin{gather}\label{equ:shortcut-pred}
    \beta_{ui} = s_{b}(\Mat{b}_{u},\Mat{b}_{i}),
\end{gather}
where $\beta_{ui}\in [0,1]$ denotes the probability of user $u$ adopting item $i$, based purely on the popularity-relevant information; $s_{b}(\cdot)$ is defined similarly to $s(\cdot)$ in Equation \eqref{equ:identity-predict}.
It is capable of quantifying the effect of the shortcut representations on predicting user preference.

Hereafter, we set softmax loss as the learning objective and minimize it to train the parameters of the shortcut model:
\begin{gather}\label{equ:shortcut-loss}
    \min_{\Theta_{b}}\Lapl_{b} = -\sum_{(u,i)\in\Set{O}^{+}}\log\frac{\exp{(\beta_{ui}/\tau)}}{\sum_{j\in\Set{N}^{+}_{u}}\exp{(\beta_{uj}/\tau)}},
\end{gather}
where $\beta_{uj}$ results from $b'=(d_{u},d_{j})$.
This optimization enforces the shortcut degree $\beta_{ui}$ to reconstruct the historical interaction, so as to distill useful information --- assessing how informative the shortcut representations $\Mat{b}_{u}$ and $\Mat{b}_{i}$ are to predict the interaction.
For example, the high value of $\beta_{ui}$ suggests the powerful predictive ability of the popularity shortcut, which easily disturbs the learning of CF models;
meanwhile, the low value of $\beta_{ui}$ shows this popularity shortcut is error-prune to predict user preference.

\subsubsection{\textbf{Overall Debiasing}}
Once the shortcut model is trained, we move forward to debiasing the target CF model.
Our intuition is that the debiased model should learn the information beyond those contained in the shortcut model.
Specifically, if the shortcut model has already achieved good prediction on a user-item pair, the debiased model should learn beyond the interaction-wise shortcut to preserve the performance;
otherwise, it needs to pursue better representations for one observed interaction with a low shortcut degree.
In a nutshell, PopGo aims to encourage the target CF model to reduce the reliance on popularity shortcuts and focus more on shortcut-agnostic information.

To this end, given a user-item pair $(u,i)$, we use its interaction-wise shortcut degree $\beta^{*}_{ui}$ as the mask of the prediction $\alpha_{ui}$, formally $\alpha_{ui}\cdot\beta^{*}_{ui}$.
On the top of the masked prediction, we minimize the following learning objective to optimize the target CF model, rather than the original loss in Equation \eqref{equ:identity-loss}:
\begin{gather}\label{equ:ours-loss}
    \min_{\Theta}\Lapl = -\sum_{(u,i)\in\Set{O}^{+}}\log\frac{\exp{(\alpha_{ui}\cdot\beta^{*}_{ui}/\tau)}}{\sum_{j\in\Set{N}^{+}_{u}}\exp{(\alpha_{uj}\cdot\beta^{*}_{uj}/\tau)}},
\end{gather}
where $\beta^{*}_{ui}$ is the outcome of the well-trained shortcut model.
% We optimize the target CF model in this way to prevent it from learning popularity shortcut.
To better interpret the impact of Equation \eqref{equ:ours-loss} on both representation learning and interaction modeling, we take two user-item pairs, $(u,i)$ and $(u,i')$, as an example.
Assuming that $\beta^{*}_{ui}\approx 1$ and $\beta^{*}_{ui'}\approx 0.2$, $\alpha_{ui}\cdot\beta^{*}_{ui}$ can easily achieve lower loss than $\alpha_{ui'}\cdot\beta^{*}_{ui'}$, thus making $(u,i)$ and $(u,i')$ function as easy and hard instances respectively.
To further reduce the losses, $\alpha_{ui'}$ needs to pursue higher scores than $\alpha_{ui}$, so as to uncover the popularity-agnostic information.
Furthermore, such hard instances offer informative gradients to guide the representation learning towards more robust representations.

The overall framework of PopGo can be summarized as follows:
% rather than adopting the original learning objective in Equation \eqref{equ:identity-loss}, we exploit the PopGo strategy to 
\begin{gather}
    \min_{\Theta}\min_{\Theta_{b}}\Lapl + \Lapl_{b}.
\end{gather}
When the target CF model is trained, we simply disable the shortcut model and deploy $\alpha_{ui}$ for the debiased prediction during inference.

%% file: chapters/4_justification.tex
\section{Justification}
Here we take both causal and information-theoretical look at PopGo.
% to justify why it is able to capture the critical popularity-agnostic information and achieve better generalization ability.

\subsection{Causal Look at PopGo}
Throughout the section, an upper-case letter (\eg $X$) represents a random variable, while its lower-cased letter (\eg $x$) denotes its observed value. 

% \vspace{5pt}
% \noindent\textbf{Causal Graph.}
\subsubsection{\textbf{Causal Graph}}
\begin{figure}[t]
    \centering
    \includegraphics[width=0.8\linewidth]{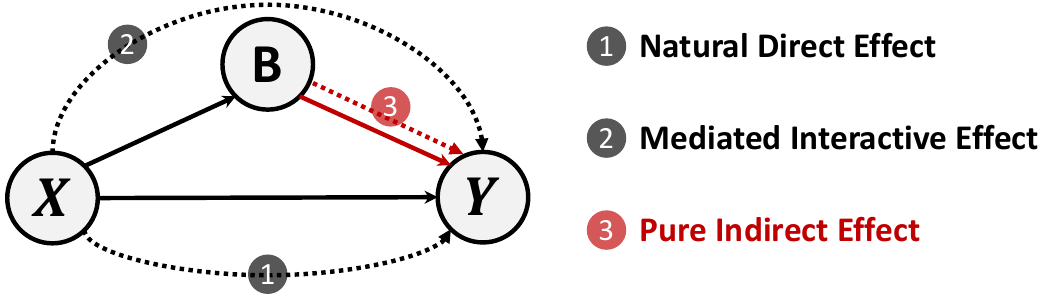}
    % \vspace{-5pt}
    \caption{Causal graph of CF, where $X$, $B$, and $Y$ denote the historical user-item interaction, popularity distribution, interaction label, respectively. \za{Considering a specific example of a user $u$ and an item $i$, we can specialize these variables to certain values: $X=(u,i)$ depicting the user and item representations, $B=(p_{u},p_{i})$ summarizing the information about the popularity distribution, $Y=y_{ui}$ indicating how likely $u$ will interact with $i$.}}
    \label{fig:causal_graph}
    % \vspace{-5pt}
\end{figure}

Following causal theory \cite{pearl2000causality,pearl2016causal}, we formalize the causal graph of CF in Figure \ref{fig:causal_graph}, which is consistent with causal graphs summarized in \cite{Bias_survey}.
It presents the causalities among three variables: user-item pair $X$, popularity feature $B$, and interaction label $Y$.
Each link represents a cause-and-effect relationship between two variables:
\begin{itemize}[leftmargin=*]
    \item $X\rightarrow B$. The popularity feature $B$ is determined by the user-item pair $X$, which consists of the user popularity and item popularity.
    \item $X\rightarrow Y\leftarrow B$. The interaction label $Y$ is made based on both user-item pair $X$, and its popularity feature $B$.
\end{itemize}
% \vspace{5pt}
% \noindent\textbf{Factual \& Counterfactual World.}
Based on the counterfactual notations \cite{pearl2000causality,pearl2016causal}, 
if $X$ is set as $x$, the value of $B$ is denoted by:
\begin{gather}
    B_{x} = B(X=x).
\end{gather}

For the treatment notation, when setting $X$ on the user-item pair $x=(u,i)$, we have $B_{x}=b=(p_{u},p_{i})$ by looking up its popularity feature.
Under no-treatment condition, $B_{x^{*}}$ describes the situation where $X$ is intervened as $x^{*}=\emptyset$.
Analogously, if $X$ and $B$ is set as $x$ and $b$ respectively, the value of $Y$ would be represented as:
\begin{gather}
    Y_{x,b} = Y(X=x,B=b).
\end{gather}
In the factual world, we have $Y_{x,B_{x}}$ by setting $X$ as $x=(u,i)$ and naturally obtaining $B_{x}=b=(p_{u},p_{i})$. 
In the counterfactual world, $Y_{x,B_{x^{*}}}=Y(X=x,B=B(X=x^{*}))$ presents the status where $X$ is set as $x$ and $B$ is set to the value when $X$ had been $x^{*}$.
Similarly, $Y_{x^{*},B_{x}}=Y(X=x^{*},B=B(X=x))$.

% \vspace{5pt}
% \noindent\textbf{Total Effect vs. Total Direct Effect.}
\subsubsection{\textbf{Total Effect vs. Total Direct Effect}}
Causal effect \cite{pearl2000causality,pearl2016causal} is the difference between two potential outcomes, when performing two different treatments.
Considering that $X=x$ is under treatment, while $X=x^{*}$ is under control, the total effect (TE) of $X=x$ on $Y$ is defined as:
\begin{gather}
    \text{TE} = Y_{x,B_{x}} - Y_{x^{*},B_{x^{*}}} = Y_{x,b} - Y_{x^{*},b^{*}}.
\end{gather}
As such, the conventional learning objective (\cf Equation \eqref{equ:identity-loss}) in essence maximizes the TE of $X=x$ on $Y$.

TE consists of total direct effect (TDE) and pure indirect effect (PIE) \cite{vanderweele2013three}, where TDE is composed of mediated interactive effect (MIE) \cite{vanderweele2013three} and natural direct effect (NDE) \cite{robins1992identifiability}.
Figure \ref{fig:causal_graph} illustrates the paths corresponding to different components.
Formally, the PIE of $X$ on $Y$ is defined as:
\begin{gather}
    \text{PIE} = Y_{x^{*},B_{x}} - Y_{x^{*},B_{x^{*}}} = Y_{x^{*},b} - Y_{x^{*},b^{*}},
\end{gather}
which reflects how the interaction label changes with the difference of popularity feature solely --- under treatment (\ie $B=B_{x}=B(X=x)$) and control (\ie $B=B_{x^{*}}=B(X=x^{*})$).
Clearly, PIE only captures the influence of popularity features, while ignoring that of user-item pair.
However, as the popularity distribution easily varies in open-world scenarios, PIE fails to suggest user preference reliably.
We hence attribute the CF models' poor generalization to memorizing PIE as the shortcut.

We now inspect the TDE of $X$ on $Y$:
\begin{gather}
    \text{TDE} = Y_{x,B_{x}} - Y_{x^{*},B_{x}}=Y_{x,b} - Y_{x^{*},b},
\end{gather}
which represents how the interaction label responses with the change of user-item pair --- under treatment (\ie $X=x=(u,i)$) and control (\ie $X=x^{*}=\emptyset$).
Distinct from PIE, TDE reflects the critical information inherent in the user-item pair, which is useful to estimate user preference.
It is worth mentioning that, as a part of TDE, MIE contains a good impact of popularity, rather than discarding all information relevant to popularity.

% \vspace{5pt}
% \noindent\textbf{Parameterization.}
\subsubsection{\textbf{Parameterization \& Training}}
Therefore, we expect the target CF models to exclude PIE and emphasize TDE.
However, it is hard to optimize TDE directly.
Here we parameterize the estimation of $Y$ as a log-likelihood of the combination of components:
\begin{gather}\label{equ:causal-overall-pred}
    Y(X,B) = \log{(Z_{X}\cdot Z_{B})},
\end{gather}
where $Z_{B}$ is the shortcut model that only takes the popularity-only features, and $Z_{X}$ is the target CF model founded upon the user-item pairs and supposes to make unbiased predictions with debiased representations.
Wherein, $Z_{B}$ is represented as:
\begin{gather}\label{equ:causal-shortcut-pred}
    Z_{B}=
    \begin{cases}
    z_{b}=\beta_{ui}, & \text{if}~~B=b\\
    z^{*}_{b}=c_{1}, & \text{if}~~B=\emptyset
    \end{cases},
\end{gather}
where $B=b = (p_{u},p_{i})$ is the treatment condition, while $B=b^{*}=\emptyset$ is the control condition.
As the shortcut model cannot deal with the invalid input $\emptyset$, we assume that it will return a constant $c_{1}$ as the probability.
In contrast, under condition of treatment $B=b$, the shortcut model will yield the probability $\beta_{ui}$ (\cf Equation \eqref{equ:shortcut-pred}).
$Z_{X}$ is represented as:
\begin{gather}\label{equ:causal-identity-pred}
    Z_{X}=
    \begin{cases}
    z_{x}=\alpha_{ui}, & \text{if}~~X=x\\
    z^{*}_{x}=c_{2}, & \text{if}~~X=\emptyset
    \end{cases},
\end{gather}
where $X=x=(u,i)$ and $X=x^{*}=\emptyset$ indicate $X$ under treatment and control, respectively.
When $X=x$, the CF model generates the probability $\alpha_{ui}$ (\cf Equation \eqref{equ:identity-predict}); meanwhile, the invalid input $\emptyset$ is assigned with the constant probability $c_{2}$.

According to Equations \eqref{equ:causal-overall-pred}, \eqref{equ:causal-shortcut-pred}, and \eqref{equ:causal-identity-pred}, we can quantify each term within TE, PIE, and TDE, such as $Y_{x,b} = \log{(z_{x}\cdot z_{b})}$.
However, directly maximizing TDE is hard to harness the CF model to disentangle the popularity-relevant and popularity-agnostic information.
Our PopGo strategy first maximizes PIE and then maximizes TE to get TDE.
Specifically, optimizing the shortcut model via Equation \eqref{equ:shortcut-loss} is consistent with learning the PIE maximum:
\begin{gather}
    \max_{\beta_{ui}}\text{PIE} = \log{(c_{2}\cdot \beta_{ui})} - \log{(c_{2}\cdot c_{1})} = \log{\beta_{ui}} - \log{c_{1}}.
\end{gather}
Once the shortcut model is well-trained, we use its outcome $\beta^{*}_{ui}$ to maximize TE, which aligns well with Equation \eqref{equ:ours-loss}:
\begin{gather}
    \max_{\alpha_{ui}}\text{TE} = \log{(\alpha_{ui}\cdot\beta^{*}_{ui})} - \log{(c_{2}\cdot c_{1})}.
\end{gather}
As a consequence, it is capable of guiding the CF models to compute TDE and get rid of PIE, which is the cause of poor generalization \wrt popularity distribution.
In summary, PopGo enforces the shortcut model and the CF model to fall into the roles of estimating PIE and TDE separately.

\subsection{Information-Theoretical Look at PopGo}
From the perspective of information theory \cite{kullback1997information}, the conventional learning paradigm of CF models (\cf Equation \eqref{equ:identity-loss}) essentially maximizes the mutual information $I(X;Y)$ between the user-item pair variable $X$ and the label variable $Y$.
However, as the popularity feature $B$ is inherent in the interaction $X$ --- that is, $B$ is a descendant of $X$ in Figure \ref{fig:causal_graph}, maximizing $I(X;Y)$ fails to shield the CF models from the popularity shortcut.

We will show that PopGo maximizes the conditional mutual information $I(X;Y|B)$ instead.
Intuitively, conditioning on a variable means fixing the variations of this variable, and hence its effect can be removed \cite{kullback1997information}.
Hence, $I(X;Y|B)$ measures the information amount contained in $X$ to predict $Y$, when conditioning on $B$ and removing its effect.
Towards this end, we elaborate on the purpose of each step in PopGo.

First, to optimize the shortcut model, PopGo minimizes Equation \eqref{equ:shortcut-loss} which is equivalent to maximizing the mutual information between the popularity variable $B$ and the interaction label variable $Y$.
Formally, negative $\Lapl_{b}$ in Equation \eqref{equ:shortcut-loss} is a lower bound of $I(B;Y)$:
\begin{gather}
    -\Lapl_{b} = - H(Y|B) \leq I(B;Y),
\end{gather}
% \begin{gather}
%     -\Lapl_{b} = H(h)\leq I(B;Y),
% \end{gather}
% \begin{gather}
%     -\Lapl_{b} = \sum_{(u,i)\in\Set{O}^{+}}\log\frac{\exp{(\beta_{ui}/\tau)}}{\sum_{j\in\Set{N}^{+}_{u}}\exp{(\beta_{uj}/\tau)}} \leq I(B;Y),
% \end{gather}
where $I(B;Y)=H(Y)-H(Y|B)$ quantifies the information amount of $Y$ by observing $B$ solely; $H(Y|B)$ is the conditional entropy, and $H(Y)$ is the marginal entropy.
The optimal critic of minimizing $\Lapl_{b}$, $\beta^{*}_{ui}$, is a log-likelihood \cite{PooleOOAT19}: $\beta^{*}_{ui} = \log{(p(y=1|b))} + c$, where $c$ is the constant.

Hereafter, to optimize the target CF model, PopGo minimizes Equation \eqref{equ:ours-loss}, which is consistent with maximizing the mutual information between $(X,B)$ and $Y$.
Analogously, $-\Lapl$ of Equation \eqref{equ:ours-loss} serves as a lower bound of $I(X,B;Y)$:
\begin{gather}
    -\Lapl = - H(Y|X,B) \leq I(X,B;Y),
\end{gather}
where $I(X,B;Y)$ assesses the amount of information about $Y$ by observing $X$ and $B$ jointly.
When using $\beta^{*}_{ui} = \log{(p(y=1|b))} + c$ and $\frac{1}{\Set{N}^{+}_{u}}\sum_{j\in\Set{N}^{+}_{u}}p(y=1|b')$ to estimate the marginal distribution $p(y=1)=\int p(b')p(y=1|b') db'$, we further rewrite $-\Lapl$ as follows:
\begin{align}
    -\Lapl&\approx \sum_{(u,i)\in\Set{O}^{+}}\log{\frac{\exp{(\alpha_{ui}/\tau)}}{\sum_{j\in\Set{N}^{+}_{u}}p(y=1|b')\exp{(\alpha_{uj}/\tau)}}}\nonumber\\
    &+\sum_{(u,i)\in\Set{O}^{+}}\log{\frac{p(y=1|b)}{p(y=1)}},
\end{align}
% \begin{align}
%     -\Lapl
%     &=\sum_{(u,i)\in\Set{O}^{+}}\log\frac{p(y=1|b)\exp{(\alpha_{ui}/\tau)}}{\sum_{j\in\Set{N}^{+}_{u}}p(y=1|b')\exp{(\alpha_{uj}/\tau)}}\\
%     &\approx \sum_{(u,i)\in\Set{O}^{+}}\log{\frac{\exp{\alpha_{ui}/\tau}}{\sum_{j\in\Set{N}^{+}_{u}}p(y=1|b')\exp{(\alpha_{uj}/\tau)}}}\\
%     &+\sum_{(u,i)\in\Set{O}^{+}}\log{\frac{p(y=1|b)}{p(y=1)}}
% \end{align}
where the second term is the estimation of $I(B;Y)$.
As $I(X,B;Y)=I(X;Y|B) + I(B;Y)$ based on the information theory, the first term approaches the desired conditional mutual information $I(X;Y|B)$. In summary, PopGo encourages the shortcut model and the target CF model to maximize $I(B;Y)$ and $I(X;Y|B)$, respectively.

%% file: chapters/5_experiments.tex
\section{Experiments}
In this section, we provide empirical results to demonstrate the effectiveness of PopGo \wx{and answer the following research questions:}
% . The experiments are designed to answer the following research questions:
\begin{itemize}
   \item \textbf{RQ1:} How does PopGo perform \wx{compared with} the state-of-the-art debiasing strategies in both OOD and ID test evaluations?
    \item \textbf{RQ2:} What are the impacts of the components (\eg the value of temperature parameter, the shortcut model) on \wx{PopGo?}
    % the improvements of PopGo's performance? Particularly,
    Can PopGo truly pursue high-quality debiased representations?
\end{itemize}

\subsection{Experimental Settings}
\textbf{Datasets.} 
\rza{We conduct experiments on six real-world benchmark datasets: Yelp2018 \cite{LightGCN}, Tencent \cite{Tencent}, Amazon-Book \cite{Amazon-Book}, Alibaba-iFashion \cite{Alibaba-ifashion}, Douban Movie \cite{Douban}, and Yahoo!R3 \cite{Yahoo}.}
To ensure the data quality, we adopt the 10-core setting \cite{Amazon-Book}, where each user and item have at least ten interaction records.
Table \ref{tab:dataset-statistics} summarizes the detailed dataset statistics.

\vspace{5pt}
\noindent\textbf{Data Splits for ID and OOD Evaluations.}
Typically, each dataset is partitioned into three parts: training, validation, and test sets.
The existing debiasing methods \cite{Causal_inference,MACR,DICE} mostly assume that either the test distribution is known in advance or the validation and test sets have the same or similar distribution, which is at odds with the training set.
Such assumptions cause the leakage of OOD test information during training the model or tuning the hyperparameters in the validation set.
However, the OOD distribution is usually unknown in the wild and open-world scenarios.
To make the evaluation more practical, we follow the standard settings of recent generalization studies \cite{DBLP:conf/nips/TeneyAKSKH20,Introspective} and create test evaluations on ID and OOD settings on Yelp2018, Tencent, Amazon-Book, and Alibaba-iFashion Datasets.
Specifically, we split each dataset into four parts: \textbf{training}, \textbf{ID validation}, \textbf{ID test}, and \textbf{OOD test} sets.
To build the OOD test set, we randomly sample $20\%$ of interactions with equal probability \wrt items, which are regarded as the most extreme balanced recommendation under an entirely random policy \cite{Causal_inference, MACR, IPS}.
We then randomly split the remaining interactions into $50\%$ for training, $10\%$ for ID validation, and $20\%$ for ID test sets.
As such, the uniform distribution in the OOD test differs from the long-tailed distribution in training, ID validation, and ID test sets.
Furthermore, we report the KL-divergence between the item popularity distribution of each set and the uniform distribution in Table \ref{tab:dataset-statistics}.
A larger KL-divergence value indicates that the heavier portion of interactions concentrates on the head of distribution. 
% See Figures \ref{fig:hist_ifashion}, \ref{fig:hist_amazon}, Table \ref{tab:popularity-statistics} in Appendix \ref{app:data-split} for more comparisons among these sets.

\vspace{5pt}
\noindent\textbf{Temporal Data Splits.}
In many applications, the distribution of popularity changes over time.
To evaluate the generalization of PopGo \wrt temporal distribution shift, we also employ the standard temporal splitting strategy \cite{Regularized_Optimization} on Douban Movie \cite{Douban} and chronologically slice the user-item interactions into the training, validation, and test sets (70\%:10\%:20\%) based on the timestamps.
\vspace{5pt}
\rza{
\noindent\textbf{Unbiased Data Splits.}
The inherent missing-not-at-random condition in real-world recommender systems makes offline evaluation of collaborative filtering a recognized challenge. 
To address this, the Yahoo!R3 \cite{Yahoo}, providing unbiased test sets collected following the missing-complete-at-random (MCAR) principle, is widely employed.
Specifically, the training data of Yahoo!R3, typically biased, includes user-selected item ratings. 
In contrast, the testing data is gathered from online surveys where users rate items chosen at random.}

% \begin{figure*}
%     \centering
%     \includegraphics[width=\linewidth]{charts/stats_yelp2018.pdf}
%     % \vspace{-10pt}
%     \caption{Histogram of the Training, Validation, OOD testing, and ID testing sets for Yelp2018.}
%     \label{fig:hist_amazon}
%     % \vspace{-10pt}
% \end{figure*}

\begin{figure*}
    \centering
    \subcaptionbox{Yelp2018.\label{fig:hist_yelp2018}}{\includegraphics[width=\linewidth]{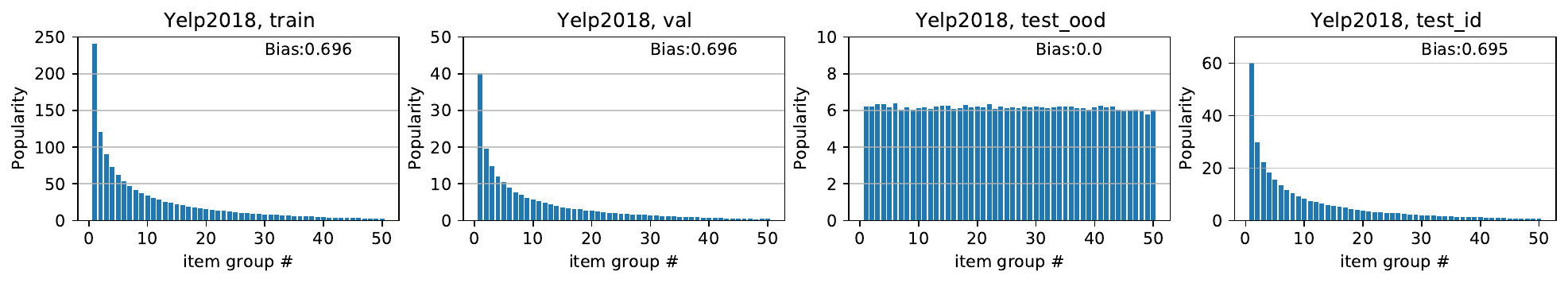}}
    \subcaptionbox{Amazon-Book.\label{fig:hist_amazon}}{\includegraphics[width=\linewidth]{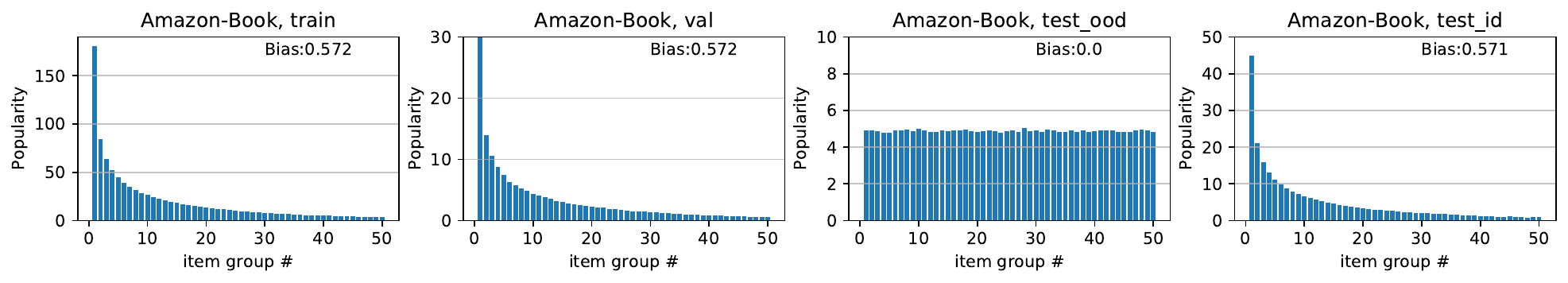}}
    \subcaptionbox{Alibaba-iFashion.\label{fig:hist_ifashion}}{\includegraphics[width=\linewidth]{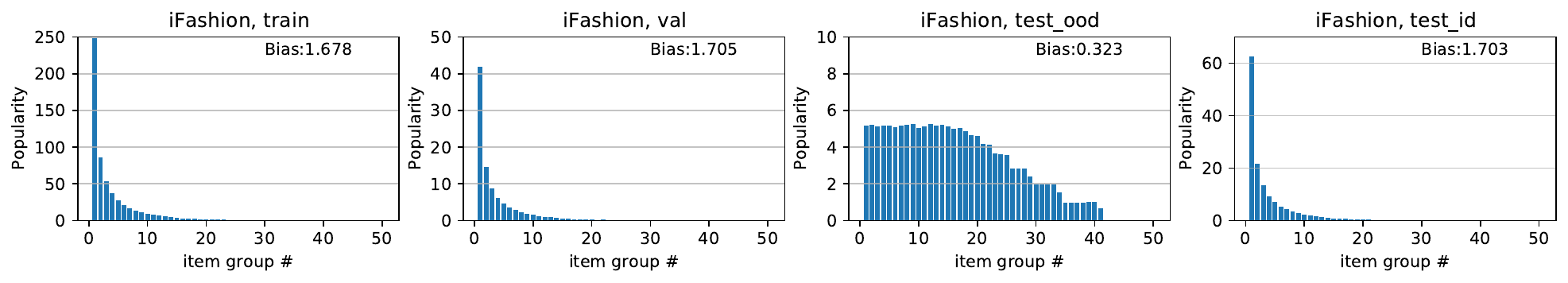}}
    \caption{Histogram of the Training, Validation, OOD testing, and ID testing sets for the Yelp2018, Amazon-Book, and Alibaba-iFashion datasets.}
    \label{fig:hist_amazon}
    % \vspace{-10pt}
\end{figure*}

% In real applications, popularity bias dynamically changes over time.
% For a comprehensive comparison and further reflecting the recommendation quality, we also conduct test splitting on Douban Movie \cite{Douban} over time. 
% Following the standard temporal splitting strategy \cite{Regularized_Optimization}, we slice the historical interactions into the training, validation, and test sets (70\%:10\%:20\%) according to the timestamps. 

% \textbf{Dataset Description.}
% To evaluate the effectiveness of PopGo, we conduct experiments on four real-world benchmark datasets: Yelp2018 \cite{LightGCN}, Tencent \cite{Tencent}, Amazon-Book \cite{Amazon-Book}, and Alibaba-iFashion \cite{Alibaba-ifashion}. 
% In each dataset, we filter out the cold-start items to ensure that each user and item have at least ten interactions. 
% All the datasets are publicly accessible and vary in terms of domain, size, sparsity, and long-tail divergence. 
% Here the long-tail divergence is defined as a KL divergence between the data distribution and uniform distribution $D_{KL}(\hat{P}_{data}|| \text{Uniform} )$, in order to show how heavy the portion of interactions concentrates on the head of distribution.
% The detailed statistics of these datasets are summarized in Table \ref{tab:dataset-statistics}. 

\begin{table}[t]
\centering
\caption{Dataset statistics.}
% \vspace{-5pt}
\label{tab:dataset-statistics}
\resizebox{0.75\linewidth}{!}{
\begin{tabular}{lrrrrrr}
\hline
 & Douban & Yelp2018 & Tencent & Amazon-Book & iFashion & Yahoo!R3 \\ \hline\hline
\#Users & 36,644 & 31,688 & 95,709 & 52,643 & 300,000 & 14,382 \\
\#Items & 22,226 & 38,048 & 41,602 & 91,599 & 81,614  & 1,000\\
\#Interactions & 5,397,926 & 1,561,406 & 2,937,228 & 2,984,108 & 1,607,813 & 129,748 \\
Sparsity & 0.00663 & 0.00130 & 0.00074 & 0.00062 & 0.00007 & 0.00902\\ \hline
$D_{KL}$-Train & 1.471 & 0.713 & 1.735 & 0.684 & 2.119 & 0.854 \\
$D_{KL}$-Validation & 1.642 & 0.713 & 1.733 & 0.684 & 2.123 & 0.822 \\
$D_{KL}$-Test\_OOD & - & 0.000 & 0.015 & 0.000 & 0.435 & - \\
$D_{KL}$-Test\_ID & - & 0.713 & 1.734 & 0.683 & 2.121 & - \\ 
$D_{KL}$-Test\_temporal & 1.428 & - & - & - & - & -   \\
$D_{KL}$-Test\_unbiased & - & - & - & - & - &  0.002\\
\hline
\end{tabular}}
% \vspace{-5pt}
\end{table}

\begin{table}[ht]
\centering
\caption{Additional Parameter statistics of PopGo.}
% \vspace{-5pt}
\label{tab:parameter-statistics}
\resizebox{0.75\linewidth}{!}{
\begin{tabular}{lrrrrr}
\hline
 & Yelp2018 & Tencent & Amazon-Book & Alibaba-iFashion & Yahoo!R3 \\ \hline\hline
User Pop Embed. &450*64 & 327*64 & 72*64 & 94*64 & 82*64 \\
Item Pop Embed. &415*64 & 396*64 & 882*64 & 545*64 & 239*64 \\
\hline
\end{tabular}}
% \vspace{-5pt}
\end{table}

\vspace{5pt}
\noindent\textbf{Evaluation Metrics.}
% In the testing phase, we conduct the all-ranking strategy \cite{KricheneR20} --- that is, for each user, all items are ranked by the CF models --- and then report three widely-used metrics: HR@$K$, Recall@$K$, NDCG@$K$, where $K$ is set as $20$ by default.
% See Appendix \ref{app:evaluation-metrics} for more details.
In the evaluation phase, we conduct the all-ranking strategy \cite{KricheneR20}, rather than sampling a subset of users \cite{WangZZLZLW19,WangLTCL20}. More specifically, each observed user-item interaction is a positive instance, while an item that the user did not adopt before is randomly sampled to pair the user as a negative instance. All these items are ranked based on the predictions of the recommender model. To evaluate top recommendation, three widely used metrics \cite{LightGCN,KricheneR20,NCF,MACR,DICE} are reported by averaging: Hit Ratio (HR), Recall and Normalized Discounted Cumulative Gain (NDCG) cut at $K$, where $K$ is set as 20 by default.

% \textbf{Evaluation Metrics.}
% In the evaluation phase, we conduct the all-ranking strategy \cite{KricheneR20}, rather than sampling a subset of users \cite{WangZZLZLW19,WangLTCL20}. More specifically, each observed user-item interaction is a positive instance, while an item that the user did not adopt before is randomly sampled to pair the user as a negative instance. All these items are ranked based on the predictions of the recommender model. To evaluate top recommendation, three widely used metrics are reported by averaging: Hit Ratio (HR), Recall and Normalized Discounted Cumulative Gain (NDCG) cut at $K$, where $K$ is set as 20 by default.

\vspace{5pt}
\noindent\textbf{Baselines.}
We select two high-performing CF models, matrix factorization (MF) \cite{BPR,MF} and LightGCN \cite{LightGCN}, to debias, which are the representative of conventional and state-of-the-art backbone models.
\rza{To demonstrate the broad applicability and superior performance of our proposed PopGo, we conduct additional experiments using the popular CF backbone - UltraGCN \cite{UltraGCN}.}
% as the target backbones being debiased, 
% We compare PopGo with the state-of-the-art debiasing strategies, covering IPS-CN \cite{IPS-CN}, CausE \cite{CausE}, DICE \cite{DICE}, and MACR \cite{MACR}.
% See Appendix \ref{app:baselines} for the introduction of each target model and debiasing strategies
For adequate comparisons, we compare PopGo with high-performing in-processing debiasing strategies in almost all research directions, including Inverse Propensity Score method (IPS-CN \cite{IPS-CN}), domain adaption method (CausE \cite{CausE}), causal embedding method (DICE \cite{DICE}, MACR \cite{MACR}), and regularization-based method (SAM-REG \cite{Regularized_Optimization}).
\begin{itemize}
    \item \textbf{MF} \cite{BPR}: Matrix Factorization (MF) is a classical CF method that uses user and item identity representations to predict, optimized by the Bayesian Personalized Ranking (BPR) loss.
    \item \textbf{LightGCN} \cite{LightGCN}: LightGCN is the state-of-the-art CF model that learns user and item representations by linearly propagating embeddings on the user-item interaction graph and weighted summing them as the final representations.
    \rza{\item \textbf{UltraGCN} \cite{UltraGCN}: UltraGCN is an ultra-simplified version of Graph Convolutional Networks (GCNs) designed for efficient recommendation systems. It skips the traditional message-passing mechanism, which can slow training, and approximates the limit of infinite-layer graph convolutions using a constraint loss. }
    \item \textbf{IPS-CN} \cite{IPS-CN}: IPS family of methods \cite{IPS,Propensity_SVM-Rank,IPS-C} eliminate popularity bias by re-weighting each instance according to item popularity. IPS-CN further adds normalization to achieve lower variance at the expense of a higher bias.
    \item \textbf{CausE} \cite{CausE}: CausE is a domain adaptation algorithm that requires one biased and one unbiased dataset to benefit its recommendation. Our experiments separate the training set into a 10\% debiased uniform training set (same as our OOD test distribution) and a 90\% biased training set. 
    \item \textbf{DICE} \cite{DICE}: Dice learns disentangled causal embeddings for interest and conformity by training with cause-specific data according to the collider effect in causal inference. 
    \item \textbf{MACR} \cite{MACR}: MACR performs counterfactual inference to remove the popularity bias by estimating and eliminating the direct effect from the item node to the prediction score.
    \rza{\item \textbf{SAM-REG} \cite{Regularized_Optimization}: SAM-REG consists of two components: the training examples mining balances the distribution of observed and unobserved items, and the regularized optimization minimizes the biased correlation between predicted user-item relevance and item popularity.}
\end{itemize}

\begin{table}[th]
% \vspace{-5pt}
\caption{Overall performance comparison in OOD test. The improvement achieved by PopGo is significant ($p$-value $<<$ 0.05).}
% \vspace{-5pt}
\label{tab:OOD}
\resizebox{0.75\linewidth}{!}{
\begin{tabular}{l|ccc|ccc}
\hline
\multicolumn{1}{c|}{OOD test} & \multicolumn{3}{c|}{Yelp2018} & \multicolumn{3}{c}{Tencent} \\
\multicolumn{1}{c|}{} & HR@20 & Recall@20 & NDCG@20 & HR@20 & Recall@20 & NDCG@20 \\\hline
ItemPop & 0.0031 & 0.0003 & 0.0007 & 0.0016 & 0.0002 & 0.0004 \\ \hline 
\textbf{MF} & 0.0558 & 0.0064 & 0.0133 & 0.0258 & 0.0047 & 0.0069 \\
+ IPS-CN & 0.0697 & \underline{0.0091} & \underline{0.0189} & 0.0359 & 0.0067 & 0.0096 \\
+ CausE & 0.0651 & 0.0073 & 0.0148 & 0.0283 & 0.0049 & 0.0070 \\
+ DICE & 0.0681 & 0.0068 & 0.0161 & 0.0380 & 0.0060 & 0.0085 \\
+ SAM-REG &   \textbf{0.0731*}  &  0.0088  &  0.0173   &   \textbf{0.0386}*   &     \underline{0.0070}   &  \underline{0.0101}\\
+ MACR & 0.0696 & 0.0076 & 0.0152 & 0.0278 & 0.0066 & 0.0077 \\
+ PopGo & \underline{0.0728} & \textbf{0.0095*} & \textbf{0.0192*} & \underline{0.0371} & \textbf{0.0072*} & \textbf{0.0103*} \\ \hline
Imp. \% & -0.41\% & 4.40\% & 1.59\% & -3.89\% & 2.86\% & 1.98\% \\ \hline
\textbf{LightGCN} & 0.0558 & 0.0069 & 0.0141 & 0.0213 & 0.0044 & 0.0054 \\
% + IPS &  &  &  &  &  &  &  &  &  &  &  &  \\ \hline
% + IPS-C &  &  &  &  &  &  &  &  &  &  &  &  \\ \hline
+ IPS-CN & 0.0664 & 0.0084 & 0.0153 & 0.0295 & 0.0063 & 0.0082\\
+ CausE & 0.0655 & 0.0082 & 0.0163 & 0.0269 & 0.0054 & 0.0064 \\
+ DICE & 0.0651 & 0.0079 & 0.0161 & 0.0414 & 0.0046 & \underline{0.0104} \\
+ SAM-REG &  \underline{0.0862}   &  \underline{0.0108}   &  \underline{0.0213}    &  \underline{0.0425}      &    0.0076   &  0.0101 \\
+ MACR & 0.0711 & 0.0092 & 0.0164 & 0.0350 & \underline{0.0081} & 0.0082 \\
+ PopGo & \textbf{0.1102*} & \textbf{0.0151*} & \textbf{0.0231*} & \textbf{0.0426*} & \textbf{0.0085*} & \textbf{0.0113*} \\\hline
Imp. \% &21.78\% & 39.81\% & 8.54\% & 0.24\% & 4.94\% & 8.65\% \\ \hline \hline
& \multicolumn{3}{c|}{Amazon-book} & \multicolumn{3}{c}{Alibaba-iFashion} \\ 
& HR@20 & Recall@20 & NDCG@20 & HR@20 & Recall@20 & NDCG@20 \\ \hline
ItemPop & 0.0013 & 0.0001 & 0.0004 & 0.0003 & 0.0008 & 0.0001 \\ \hline 
\textbf{MF} & 0.0699 & 0.0104 & 0.0195 & 0.0077 & 0.0039 & 0.0022   \\ 
+ IPS-CN  & 0.0856 & 0.0135 & 0.0237& 0.0095 & 0.0050 & 0.0027  \\ 
+ CausE & 0.0464 & 0.0050 & 0.0106 & 0.0050 & 0.0023 & 0.0012 \\
+ DICE & 0.0749 & 0.0091 & 0.0204 & 0.0088 & 0.0047 & 0.0024  \\ 
+ SAM-REG &    0.0917   &  0.0141    &     \underline{0.0279} &    \underline{0.0127} &    \underline{0.0067}  &  \underline{0.0031} \\ 
+ MACR & \underline{0.0919} & \underline{0.0155} & 0.0249 & 0.0075 & 0.0037 & 0.0016 \\ 
+ PopGo & \textbf{0.0981*} & \textbf{0.0162*} & \textbf{0.0297*} & \textbf{0.0146*}  & \textbf{0.0076*} & \textbf{0.0038*} \\ \hline
Imp. \% & 6.75\% & 4.52\% & 6.45\% & 14.96\% & 13.43\% & 22.58\% \\ \hline
\textbf{LightGCN} & 0.0782 & 0.0117 & 0.0208 & 0.0069 & 0.0033 & 0.0014 \\ 
+ IPS-CN & 0.1015 & \underline{0.0167} & \underline{0.0304} & 0.0080 & 0.0038 & 0.0017 \\
+ CausE & 0.0831 & 0.0133 & 0.0217 & 0.0063 & 0.0028 & 0.0012 \\
+ DICE & 0.0700 & 0.0086 & 0.0186 & 0.0091 & 0.0043 & 0.0020  \\
+ SAM-REG &  \underline{0.1049}   &  0.0157    &  0.0295    &  \underline{0.0110}    &    \underline{0.0050}  &  \underline{0.0028}  \\
+ MACR & 0.0990 & 0.0146 & 0.0267 & 0.0072 & 0.0037 & 0.0017 \\ 
+ PopGo & \textbf{0.1332*} & \textbf{0.0232*} & \textbf{0.0398*} & \textbf{0.0164*} & \textbf{0.0093*} & \textbf{0.0040*} \\ \hline
Imp. \% & 26.98\%  & 38.92\% & 30.92\% & 49.09\% & 86.00\% & 42.86\% \\ \hline
\end{tabular}}
% \vspace{-5pt}
\end{table}

\begin{table}[th]
\caption{Overall performance comparison in ID test. The improvement achieved by PopGo is significant ($p$-value $<<$ 0.05).}
\label{tab:ID}
% \vspace{-5pt}
\resizebox{0.75\linewidth}{!}{
% \begin{tabular}{l|ccc|ccc|ccc|ccc}
\begin{tabular}{l|ccc|ccc}
\hline
\multicolumn{1}{c|}{ID test} & \multicolumn{3}{c|}{Yelp2018} & \multicolumn{3}{c}{Tencent} \\ 
% \multicolumn{1}{c|}{} & HR & Recall & NDCG & HR & Recall & NDCG & HR & Recall & NDCG & HR & Recall & NDCG \\ \hline
\multicolumn{1}{c|}{} & HR@20 & Recall@20 & NDCG@20 & HR@20 & Recall@20 & NDCG@20 \\ \hline
ItemPop & 0.0920 & 0.0175 & 0.0184 & 0.1639 & 0.0384 & 0.0274 \\ \hline 
\textbf{MF} & \underline{0.3650} & \underline{0.0801} & \underline{0.1042} & \underline{0.2988} & \underline{0.0874} & \underline{0.0871} \\
% + IPS & 0.316 & 0.066 & 0.084 & 0.250 & 0.072 & 0.066 & 0.342 & 0.077 & 0.099 & 0.088 & 0.059 & 0.025 \\ \hline
% + IPS-C & 0.325 & 0.070 & 0.086 & 0.252 & 0.072 & 0.066 & 0.341 & 0.072 & 0.066 & 0.087 & 0.059 & 0.024 \\ \hline
+ IPS-CN & 0.3415 & 0.0735 & 0.0902 & 0.2296 & 0.0663 & 0.0574 \\
+ CausE & 0.2820 & 0.0558 & 0.0793 & 0.2410 & 0.0648 & 0.0750 \\
+ DICE & 0.1760 & 0.0281 & 0.0410 & 0.2579 & 0.0736 & 0.0782 \\
+ SAM-REG &  0.2431  &  0.0458  &  0.0622   &  0.1500    &  0.0406      &   0.0419 \\
+ MACR & 0.2090 & 0.0469 & 0.0531 & 0.1648 & 0.0479 & 0.0445 \\
+ PopGo & \textbf{0.4504*} & \textbf{0.1075*} & \textbf{0.1419*} & \textbf{0.4042*} & \textbf{0.1279*} & \textbf{0.1286*} \\ \hline 
Imp. \% & 23.40\% & 34.21\% & 36.18\% & 35.27\% & 46.31\% & 47.73\% \\ \hline
\textbf{LightGCN} & \underline{0.4124} & \underline{0.0938} & \underline{0.1210} & \underline{0.3547} & \underline{0.1067} & \underline{0.1067} \\
% + IPS &  &  &  &  &  &  &  &  &  &  &  &  \\ \hline
% + IPS-C &  &  &  &  &  &  &  &  &  &  &  &  \\ \hline
+ IPS-CN & 0.3548 & 0.0771 & 0.103 & 0.2844 & 0.0811 & 0.0878 \\
+ CausE & 0.3905 & 0.0874 & 0.1172 & 0.3294 & 0.0966 & 0.1053 \\
+ DICE & 0.1714 & 0.0335 & 0.0418 & 0.2443 & 0.0483 & 0.0683 \\
+ SAM-REG &   0.3180   &   0.0657      &    0.0893    &   0.2279     &   0.0653      &  0.0663 \\
+ MACR & 0.2650 & 0.0490 & 0.0832 & 0.2925 & 0.0830 & 0.0967 \\
+ PopGo & \textbf{0.4376*} & \textbf{0.1037*} & \textbf{0.1402*} & \textbf{0.3974*} & \textbf{0.1178*} & \textbf{0.1286*} \\ \hline
Imp. \% & 6.11\% & 10.55\% & 15.87\% & 12.04\% & 10.92\% & 20.52\% \\ \hline \hline
& \multicolumn{3}{c|}{Amazon-book} & \multicolumn{3}{c}{Alibaba-iFashion} \\
\multicolumn{1}{c|}{} & HR@20 & Recall@20 & NDCG@20 & HR@20 & Recall@20 & NDCG@20 \\ \hline
ItemPop & 0.0639 & 0.0102 & 0.0104 & 0.0348 & 0.0212 & 0.0063 \\ \hline 
\textbf{MF} & \underline{0.3621} & \underline{0.0825} & \underline{0.1102} & \underline{0.0912} & \underline{0.0603} & \underline{0.0260} \\ 
+ IPS-CN & 0.3361 & 0.0753 & 0.0970 & 0.0817 & 0.0558 & 0.0212 \\ 
+ CausE & 0.2578 & 0.0486 & 0.0736 & 0.0566 & 0.0369 & 0.0147 \\ 
+ DICE & 0.2997 & 0.0645 & 0.0902 & 0.0630 & 0.0380 & 0.0185 \\ 
+ SAM-REG &   0.2727      &    0.0599 &  0.0815     &  0.0455   &  0.0305    &  0.0126  \\
+ MACR & 0.2510 & 0.0549 & 0.0711 & 0.0862 & 0.0557 & 0.0253 \\ 
+ PopGo & \textbf{0.4433*} & \textbf{0.1081*} & \textbf{0.1481*} & \textbf{0.1540*} & \textbf{0.1050*} & \textbf{0.0475*} \\ \hline
Imp. \% & 22.42\% & 31.08\% & 34.39\% & 36.40\% & 37.98\% & 37.28\% \\ \hline 
\textbf{LightGCN} & 0.3825 & 0.0891 & 0.1224 & \underline{0.1154} & \underline{0.0754} & \underline{0.0318} \\ 
+ IPS-CN & 0.3617 & 0.0834 & 0.1149 & 0.0986 & 0.0657 & 0.0276 \\ 
+ CausE & \underline{0.3862} & \underline{0.0981} & \underline{0.1277} & 0.0421 & 0.0435 & 0.0194 \\
+ DICE & 0.2140 & 0.0450 & 0.0630 & 0.1092 & 0.0747 & 0.0315 \\
+ SAM-REG &   0.3382  &    0.0773  &  0.1080    &  0.0748   &   0.0502     &  0.0229  \\ 
+ MACR & 0.3331 & 0.0733 & 0.1082 & 0.0770 & 0.0497 & 0.0235 \\
+ PopGo & \textbf{0.4362*} & \textbf{0.1101*} & \textbf{0.1575*} & \textbf{0.1491*} & \textbf{0.1023*} & \textbf{0.0472*} \\ \hline
Imp. \% & 12.95\% & 12.23\% & 23.34\% & 29.20\% & 35.68\% & 48.43\% \\ \hline
\end{tabular}}
% \vspace{-5pt}
\end{table}

\begin{figure*}[t]
    \centering
    \subcaptionbox{Yelp2018.\label{fig:intro_yelp2018}}{\includegraphics[width=0.55\linewidth]{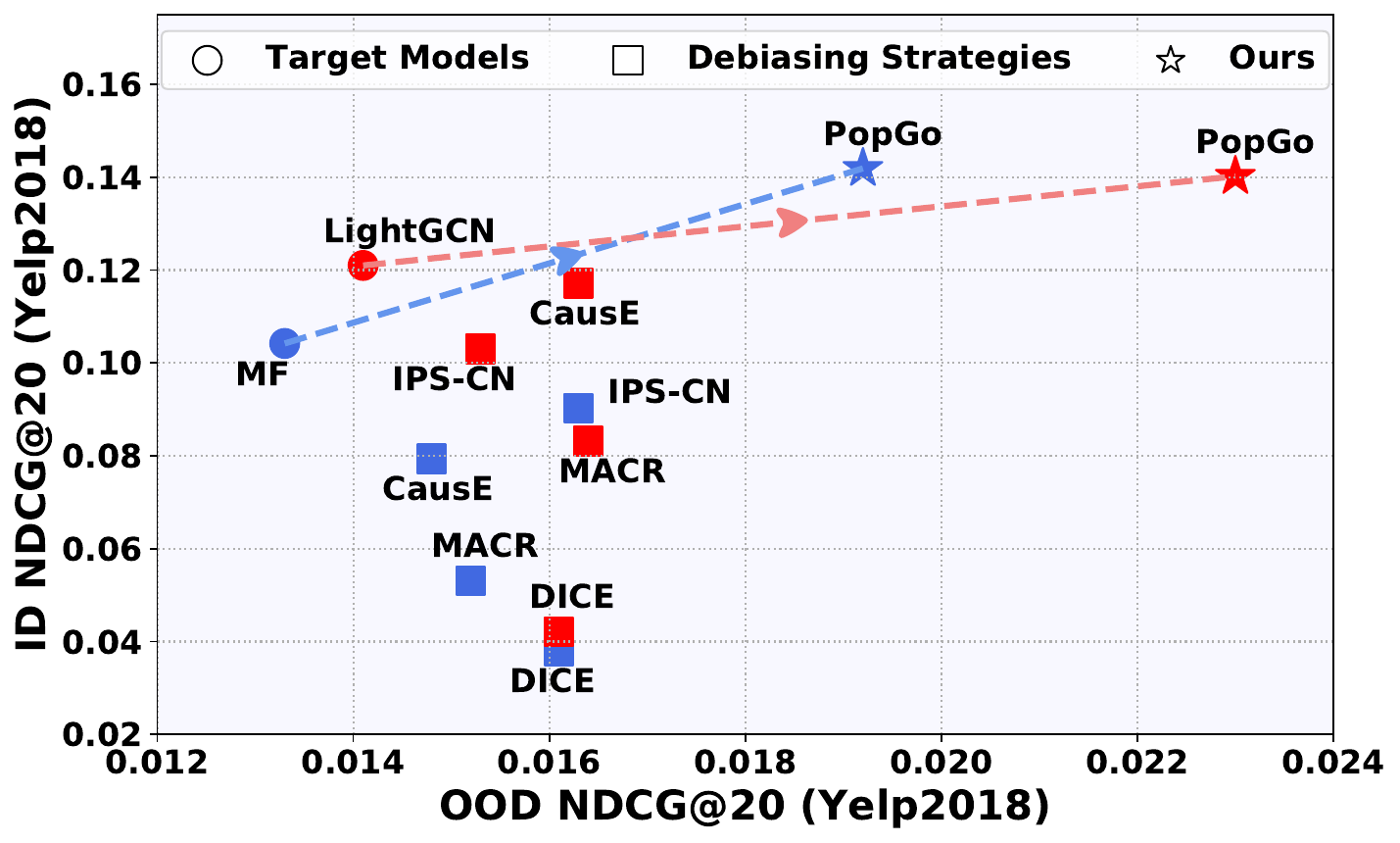}}
    \subcaptionbox{Amazon-Book.\label{fig:intro_amazon}}{\includegraphics[width=0.55\linewidth]{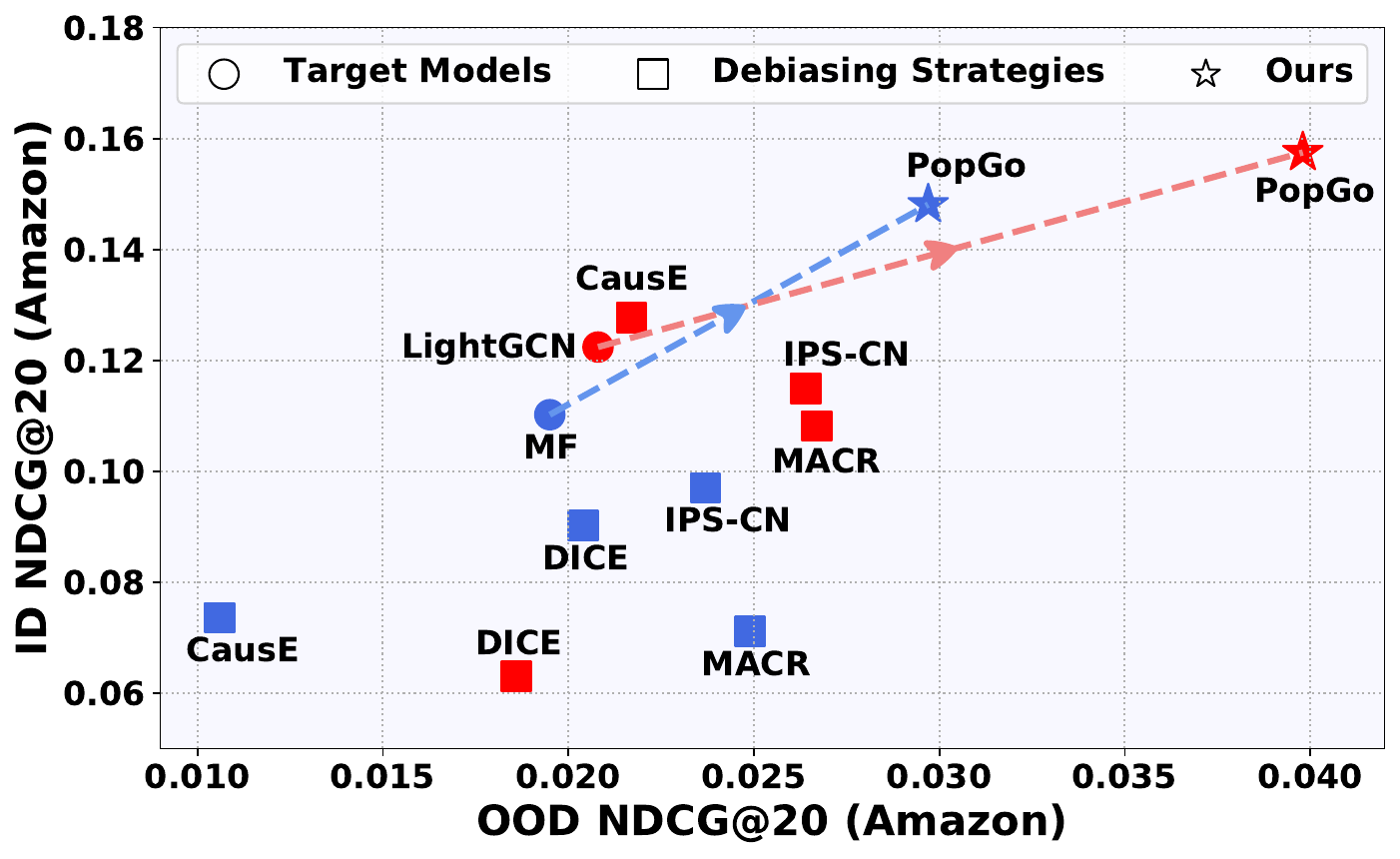}}
    \subcaptionbox{Alibaba-iFashion.\label{fig:intro_ifashion}}{\includegraphics[width=0.55\linewidth]{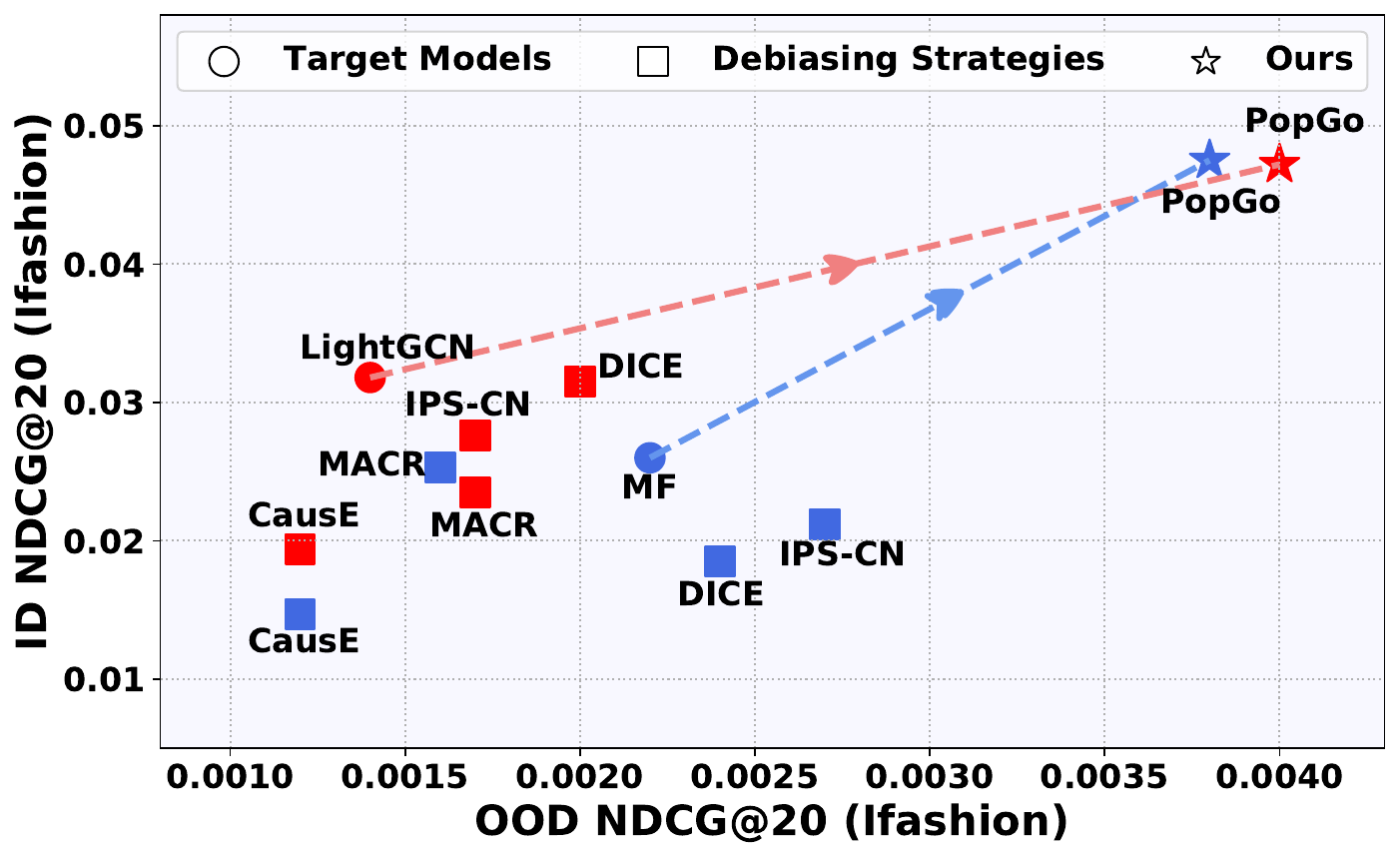}}
    \caption{Performance of debiasing strategies on the Yelp2018, Amazon-Book, and Alibaba-iFashion datasets.}
    \label{fig:intro_others}
    % \vspace{-10pt}
\end{figure*}

% \begin{figure*}[t]
%     \centering
%     \includegraphics[width=0.6\linewidth]{charts/intro_yelp2018.pdf}
%     \includegraphics[width=0.6\linewidth]{charts/intro_amazon.pdf}
%     \includegraphics[width=0.6\linewidth]{charts/intro_ifashion.pdf}
%     \caption{Performance of debiasing strategies on the Yelp2018, Amazon-Book, and Alibaba-iFashion datasets.}
%     \label{fig:intro_others}
%     % \vspace{-10pt}
% \end{figure*}

% \section{Appendix}
\subsection{Detailed Experimental Settings}
\label{app:data-split}

\vspace{5pt}
\noindent\textbf{Parameter Settings.}
We implement our PopGo in TensorFlow.
Our codes, datasets, and hyperparameter settings are available at \url{https://github.com/anzhang314/PopGo} to guarantee reproducibility.
All experiments are conducted on a single Tesla-V100 GPU.
For a fair comparison, all methods are optimized by Adam \cite{Adam} optimizer with the embedding size as 64, learning rate as 1e-3, max epoch of 400, and the coefficient of regularization as 1e-5 in all experiments.
The batch size on Yelp2018, Tencent, Amazon-Book, and Douban is 2048, while the batch size for MF-based models on iFashion is set to 256 to avoid bad convergence.
Following the default setting in \cite{LightGCN}, the number of embedding layers for LightGCN is set to 2.
We adopt the early stop strategy that stops training if Recall@20 on the validation set does not increase for 10 successive epochs.
A grid search is conducted to tune the critical hyperparameters of each strategy to choose the best models \wrt Recall@20 on the validation set.

For PopGo, we can search the $\tau$ in a wide range to obtain the best performance. But in this paper, across all datasets, we report $\tau = 0.07$. We emphasize that the performance of PopGo can be further improved through a finer grid search in the specific dataset as shown in Section \ref{subsec:tau}. The pre-training stage for the shortcut models is set to 5 epochs, then is frozen afterward. Negative sampling size is set to 64 for PopGO-MF, while inbatch-negative sampling is used for PopGO-LightGCN to reduce time for training. 
The size of additional parameters compared with MF that comes from the shortcut model is listed in Table \ref{tab:parameter-statistics}.

For IPS-CN, we follow the paper \cite{DICE} to add re-weighting factors.
For CausE, 10\% of training data with uniform distribution is used as the intervened set. The counterfactual penalty factor $cf\_pen$ is tuned in the range of $[0.01,0.1]$ and $cf\_pen=0.05$ are reported for best overall performance.
For MACR, we follow the original settings \cite{MACR} to set weights for user branch $\alpha=1e-3$ and item branch $\beta=1e-3$ respectively. We further tune hyperparameter $c=0,10,20,30,40,50$. Notice that directly tuning hyper-parameter for best validation performance(ID) leads to small $c$, which degenerates MACR to backbone model with no debiasing effect. Therefore, we manually fix $c=40$ for overall best performance. 
For SAM-REG, we follow the default setting \cite{Regularized_Optimization} by setting $rweight=0.05$.
For DICE, we set conformity loss $\alpha=0.1$ and discrepancy loss $\beta=0.01$ respectively, and set margin decay and loss decay to 0.9. The margin value is set for different datasets (10 for Amazon-book, Tencent, and Yelp2018, 2 for iFashion) according to the mean and median of item popularity for best performance.

\begin{table}[ht]
\caption{Effect of shortcut model on PopGo.}
% \vspace{-5pt}
\label{tab:ablation1} 
\resizebox{\linewidth}{!}{
\begin{tabular}{c|l|cc|cc|cc|cc}
\hline
\multicolumn{2}{l|}{\multirow{3}{*}{}} & \multicolumn{4}{c|}{Yelp2018} & \multicolumn{4}{c}{Tencent} \\ 
\cline{3-10}  
\multicolumn{2}{l|}{} & \multicolumn{2}{c|}{OOD Test} & \multicolumn{2}{c|}{ID Test} & \multicolumn{2}{c|}{OOD Test} & \multicolumn{2}{c}{ID Test} \\
\multicolumn{2}{l|}{} & Recall@20 & NDCG@20 & Recall@20 & NDCG@20 & Recall@20 & NDCG@20 & Recall@20 & NDCG@20 \\ \hline 
\multirow{3}{*}{MF}  & PopGo & 0.1050 & 0.0475  & 0.0095 & 0.0192 & 0.1075 & 0.1419 & 0.0072 & 0.0103 \\
& PopGo-S & 0.0761 & 0.0346 & 0.0081 & 0.0172 & 0.0891 & 0.1207 & 0.0054 & 0.0082 \\ \cline{2-10}
& Imp. \% & 40.0\% & 37.3\%  & 17.3\% & 11.6\% & 20.7\% & 17.56\% & 33.3\% & 25.6\% \\ \hline
\multirow{3}{*}{LightGCN}  & PopGo & 0.1023 & 0.0472& 0.0151 & 0.0230 & 0.1037 & 0.1402 & 0.0085 & 0.0113\\
& PopGo-S & 0.0704 & 0.0324 & 0.0101 & 0.0184 & 0.0800 & 0.1125 & 0.0057 & 0.0075 \\ \cline{2-10} 
& Imp. \% & 31.2\% & 33.7\% & 49.5\% & 25.0\% & 29.6\% & 24.6\% & 49.1\% & 50.7\% \\ \hline \hline
\multicolumn{2}{l|}{\multirow{3}{*}{}} & \multicolumn{4}{c|}{Amazon-Book} & \multicolumn{4}{c}{Alibaba-iFashion}\\\cline{3-10} 
\multicolumn{2}{l|}{}& \multicolumn{2}{c|}{OOD Test} & \multicolumn{2}{c|}{ID Test} & \multicolumn{2}{c|}{OOD Test} & \multicolumn{2}{c}{ID Test}\\ 
\multicolumn{2}{l|}{} & Recall@20 & NDCG@20 & Recall@20 & NDCG@20 & Recall@20 & NDCG@20 & Recall@20 & NDCG@20\\ \hline \hline
\multirow{3}{*}{MF}  & PopGo & 0.1279 & 0.1286 & 0.0162 & 0.0297 & 0.1081 & 0.1481 & 0.0076 & 0.0038\\
& PopGo-S & 0.1125 & 0.1175 & 0.0136 & 0.0268 & 0.1042 & 0.1468 & 0.0047 & 0.0024 \\ \cline{2-10} 
& Imp. \% & 13.7\% & 9.45\% & 19.1\% & 10.8\% & 3.7\% & 0.9\% & 61.7\% & 58.3\% \\ \hline 
\multirow{3}{*}{LightGCN}  & PopGo  & 0.1178 & 0.1286 & 0.0232 & 0.0398 & 0.1101 & 0.1575 & 0.0083 & 0.0040 \\ 
& PopGo-S & 0.0981 & 0.1096 & 0.0188 & 0.0320 & 0.1008 & 0.1441 & 0.0041 & 0.0022\\ \cline{2-10}
& Imp. \% & 20.1\% & 17.3\% & 23.4\% & 24.4\% & 9.2\% & 9.3\% & 97.6\% & 81.8\% \\ \hline 
\end{tabular}}
% \vspace{-5pt}
\end{table}

\begin{table}[t]
\caption{The performance comparison on Douban dataset. }
\label{tab:time_split}
% \vspace{-5pt}
\resizebox{0.75\columnwidth}{!}{
\begin{tabular}{l|ccc|ccc}
\hline
 & \multicolumn{3}{c|}{MF} & \multicolumn{3}{c}{LightGCN} \\
\multicolumn{1}{c|}{} & HR@20 & Recall@20 & NDCG@20 & HR@20 & Recall@20 & NDCG@20 \\\hline\hline
Backbone & \underline{0.2924} & \underline{0.0294} & \underline{0.0472} & \underline{0.3543} & \underline{0.0313} & \underline{0.0602} \\
+ IPS-CN & 0.2514 & 0.0174 & 0.0324 & 0.3212 & 0.0261 & 0.0502 \\
+ CausE & 0.2725 & 0.0203 & 0.0376 & 0.3403 & 0.0275 & 0.0514 \\
+ SAM-REG & 0.2826 & 0.0191 & 0.0390 & 0.2944 & 0.0252 &  0.0488 \\
+ MACR & 0.1084 & 0.0087 & 0.0163 & 0.3127 & 0.0271 & 0.0519 \\
+ PopGo & \textbf{0.3776}* & \textbf{0.0333}* & \textbf{0.0613}* & \textbf{0.3613}* & \textbf{0.0349}* & \textbf{0.0660}* \\\hline
Imp. \% & 29.1\% & 13.3\% & 29.9\% & 2.0\% & 11.5\% & 9.7\% \\\hline
\end{tabular}}
% \vspace{-5pt}
\end{table}

\begin{table}[t]
    \centering
    \caption{Performance comparison in the unbiased test under UltraGCN backbone.}
    \label{tab:yahoo}
    \resizebox{0.45\linewidth}{!}{
    \begin{tabular}{l|ccc}
    \toprule
     & \multicolumn{3}{c}{Yahoo!R3}  \\ 
     
    &HR@20 & Recall@20 & NDCG@20  \\\midrule
    \textbf{UltraGCN} & 0.1972 & 0.1309 & 0.0599  \\
    + IPS-CN & 0.1997 & 0.1315 & 0.0595  \\
    + CausE & \underline{0.2060} & \underline{0.1324} & \underline{0.0619} \\
    + SAM-REG &   0.1964  &  0.1292  &  0.0583  \\
    + MACR & 0.2023 & 0.1319 & 0.0591 \\
    + PopGo & \textbf{0.2161*} & \textbf{0.1403*} & \textbf{0.0635*} \\ \hline
    % Imp. \% & 1.78\% & -2.16\% & 1.44\% \\ \hline
    
    \end{tabular}}
  \end{table}

\begin{table}[ht]
\centering
\caption{Training cost on Tencent (seconds per epoch/total).}
% \vspace{-5pt}
\label{tab:elapse_time}
\resizebox{0.95\linewidth}{!}{
\begin{tabular}{lrrrrrrr}
\hline
 & Backbone & +IPS-CN & +CausE & +DICE& +SAM-REG & +MACR & +PopGO\\ \hline\hline
MF & 15.5/17887 & 17.8/10662 & 16.6/1859 & 66.7/38152 & 18.2/3458 & 160/17600 &  35.1/2141 \\
LightGCN & 78.6/4147 & 108/23652 & 47.2/3376 & 126/6804 & 49.8/10458 & 135/20250 &  141/3102 \\ \hline
\end{tabular}}
% \vspace{-5pt}
\end{table}

\subsection{Performance Comparison (RQ1)}
Table \ref{tab:OOD} and Table \ref{tab:ID} report the comparison of debiasing performance in OOD and ID test evaluations, respectively.
Wherein the best performing methods are bold and starred, and the strongest baselines are underlined; Imp.\% measures the relative improvements of PopGo over the strongest baselines.
We observe that:
\begin{itemize} 
    \item \textbf{In both OOD and ID evaluations, PopGo significantly outperforms the baselines across four datasets in most cases.} Specifically, it achieves remarkable improvements over the best debiasing baselines and target CF models by 1.59\% - 42.86\% and 15.87\% - 48.43\% \wrt NDCG@20 in OOD and ID settings, respectively.
    Clearly, PopGo is not only superior to the debiasing strategies in the OOD settings but also performs better than the original CF models in the ID settings. 
    This verifies that PopGo improves the generalization of backbones, rather than seeking a trade-off between ID and OOD performance.

    \item \textbf{Jointly analyzing the debiasing baselines in Tables \ref{tab:OOD}, \ref{tab:ID} and Figure \ref{fig:intro_tencent}, there is a trade-off trend between the ID and OOD performance.}
    Generally, with the increase of OOD results, their ID performance is becoming worse. We conjecture that their OOD improvements may come from the sacrifice of ID performance, thus they might generalize poorly in the wild.
    
    \item \textbf{PopGo is able to capture the model-dependent popularity bias.} Specifically, over other debiasing strategies, the improvements achieved by LightGCN+PopGo are more significant than that by MF+PopGo.
    As discussed in Section \ref{sec:cf-learning-paradigm}, LightGCN could amplify the bias during the information propagation, thus owning stronger biases than MF.
    It inspires us to reduce the potential biases caused by the model architecture.
    By cloning the target CF model as the shortcut model (\cf Section \ref{sec:shortcut-rep-learn}), our PopGo can package the model-dependent bias into the shortcut representations and represent the bias via the interaction-wise shortcut degree more accurately.
    
    \item \textbf{Debiasing baselines perform unstably across datasets.} For example, in Alibaba-iFashion, the OOD results of MF+CausE, MF+MACR, and MF+DICE are lower than MF \wrt all metrics.
    Compared with other datasets, Alibaba-iFashion is the sparsest and most biased, making it more challenging to be debiased.
    Without an effective measurement of interaction-wise shortcut degrees or the OOD test prior to painstakingly tune the hyperparameters, the debiasing ability of these baselines is limited. In contrast, benefiting from no assumption on the test data, PopGo can attain remarkable improvements ($40\%$ - $100\%$ on Alibaba-iFahsion) in the OOD setting.

    \item \textbf{PopGo can leverage the popularity information to boost the ID performance.} Interaction sparsity influences the representation quality of the target models.
    As the sparsity increases across datasets, PopGo significantly promotes the ID performance compared with the non-debiased backbones in most cases.
    Here we attribute this to (1) the MIE of popularity captured by PopGo, from the causal-theoretical view; and (2) the focus on mining the hard examples that are with sparser interactions (See the visual evidence in Figure \ref{fig:visual-embeds}). 

\end{itemize}
Moreover, Table \ref{tab:elapse_time} shows the training time of all methods, where PopGo has the comparable time complexity to the backbone models.
Furthermore, Table \ref{tab:time_split} elaborates the comparison of performance on the temporal split dataset, Douban Movie.
\rza{Table \ref{tab:yahoo} presents a comparative analysis of debiasing performance in unbiased test evaluations, Yahoo!R3.}
We observe that: 
\begin{itemize}
    \item \textbf{PopGo consistently outperforms all baselines in terms of all metrics on Douban Movie.}
    For instance, it achieves significant gains over the target models - MF and LightGCN \wrt NDCG@20 by 29.9\% and 9.7\%, respectively.
    We attribute these results to successfully improving the generalization of target CF models against the popularity distribution shift over time.
    \item \textbf{None of the debiasing baselines could maintain a comparable performance to the original CF models on temporal split settings.}
    Surprisingly, even the state-of-the-art popularity debiasing technique - MACR, fails to diagnose the popularity bias when the bias changes over time.
    We ascribe the failure to their disability of measuring instance-wise popularity bias and the trade-off of sacrificing accuracy to promote item coverage.
    \item \textbf{PopGo consistently surpasses other debiasing methods and the UltraGCN backbone model in unbiased evaluations on Yahoo!R3.} Notably, some debiasing methods, such as MACR and SAM-REG, may even reduce recommendation accuracy. This suggests that PopGo effectively mitigates popularity bias and exhibits wider applicability in real-world scenarios.
\end{itemize}

\subsection{Study on PopGo (RQ2)}

\subsubsection{\textbf{Impact of Temperature $\tau$}} \label{subsec:tau}

\begin{figure}[t]
	\centering
	\subcaptionbox{MF+PopGo\label{fig:tau_mf}}{
	    % \vspace{-5pt}
		\includegraphics[width=0.48\linewidth]{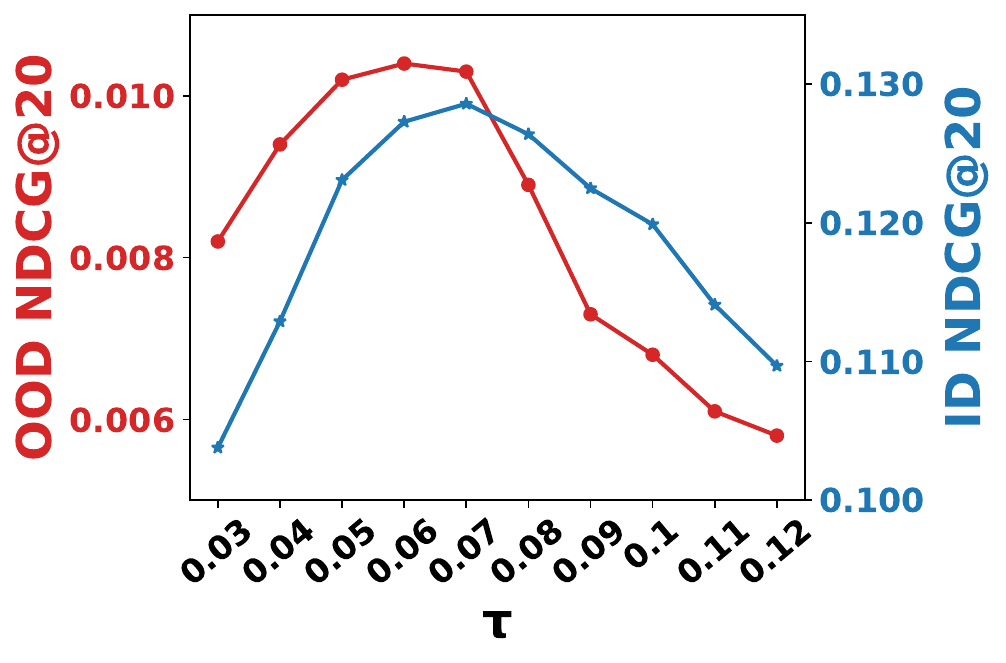}}
	\subcaptionbox{LightGCN+PopGo\label{fig:tau_Lightgcn}}{
	    % \vspace{-5pt}
		\includegraphics[width=0.48\linewidth]{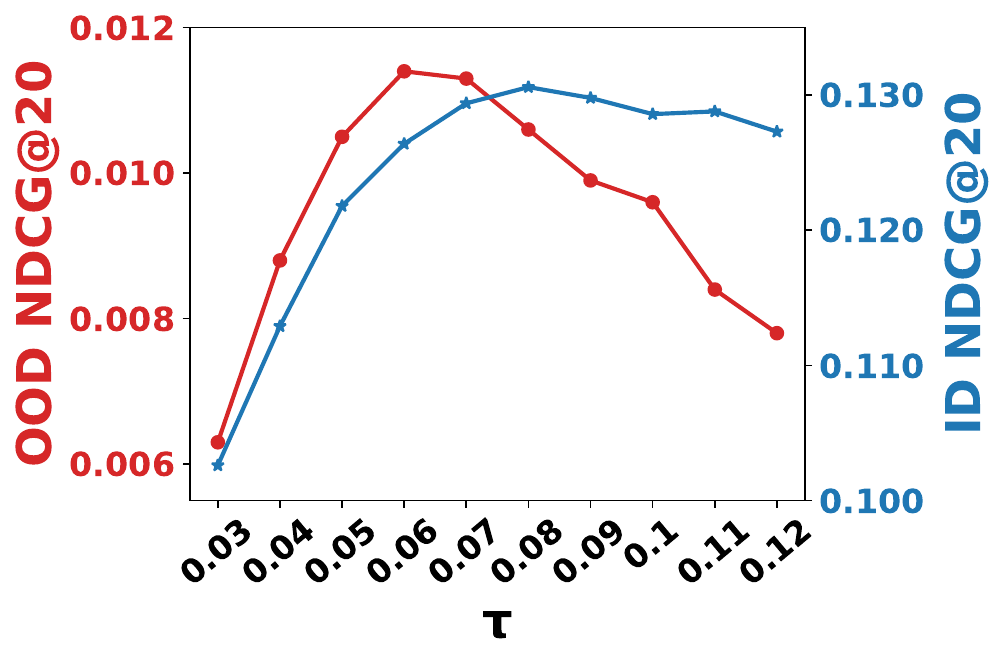}}
	% \vspace{-10pt}
	\caption{Temperature $\tau$ sensitivity analysis on Tencent.}
	\label{fig:tau}
	% \vspace{-10pt}
\end{figure}

PopGo only has one hyperparameter to tune --- temperature $\tau$ in Equation \eqref{equ:shortcut-loss} and \eqref{equ:ours-loss}.
In Figure \ref{fig:tau}, we report the sensitivity analysis of $\tau$ on Tencent and find:
\begin{itemize}
    \item Both OOD and ID evaluations exhibit the concave unimodal functions of $\tau$, where the curves reach the peak almost synchronously in a small range of $\tau$. For example, MF+PopGo gets the best performance when $\tau=0.06$ and $\tau=0.07$ in OOD and ID settings, respectively.
    This justifies that PopGo does not suffer from the trade-off between the OOD and ID evaluations, and again verifies that PopGo improvements the OOD generalization without sacrificing the ID performance.
    \item LightGCN+PopGo is more sensitive to $\tau$ than MF+PopGo in the OOD test, while being more robust in the ID test.
    We ascribe this to the graph-enhanced representations of LightGCN, which are more prone to over-emphasize popular users and items than the identity embeddings of MF (\cf Section \ref{sec:representation-learning}).
\end{itemize}

% \begin{figure}[t]
%     \centering
%     \subfigure[MF+PopGo]{\label{fig:tau_mf}
%     \includegraphics[width=0.495\linewidth]{charts/tau_MF.pdf}}
%     \subfigure[LightGCN+PopGo]{\label{fig:tau_Lightgcn}
%     \includegraphics[width=0.495\linewidth]{charts/tau_LightGCN.pdf}}
%     \caption{Temperature $\tau$ sensitivity analysis on Tencent dataset.}
%     \label{fig:tau}
% \end{figure}

\subsubsection{\textbf{Impact of Shortcut Model}} 
\begin{figure}
    \centering
    \includegraphics[width=0.65\linewidth]{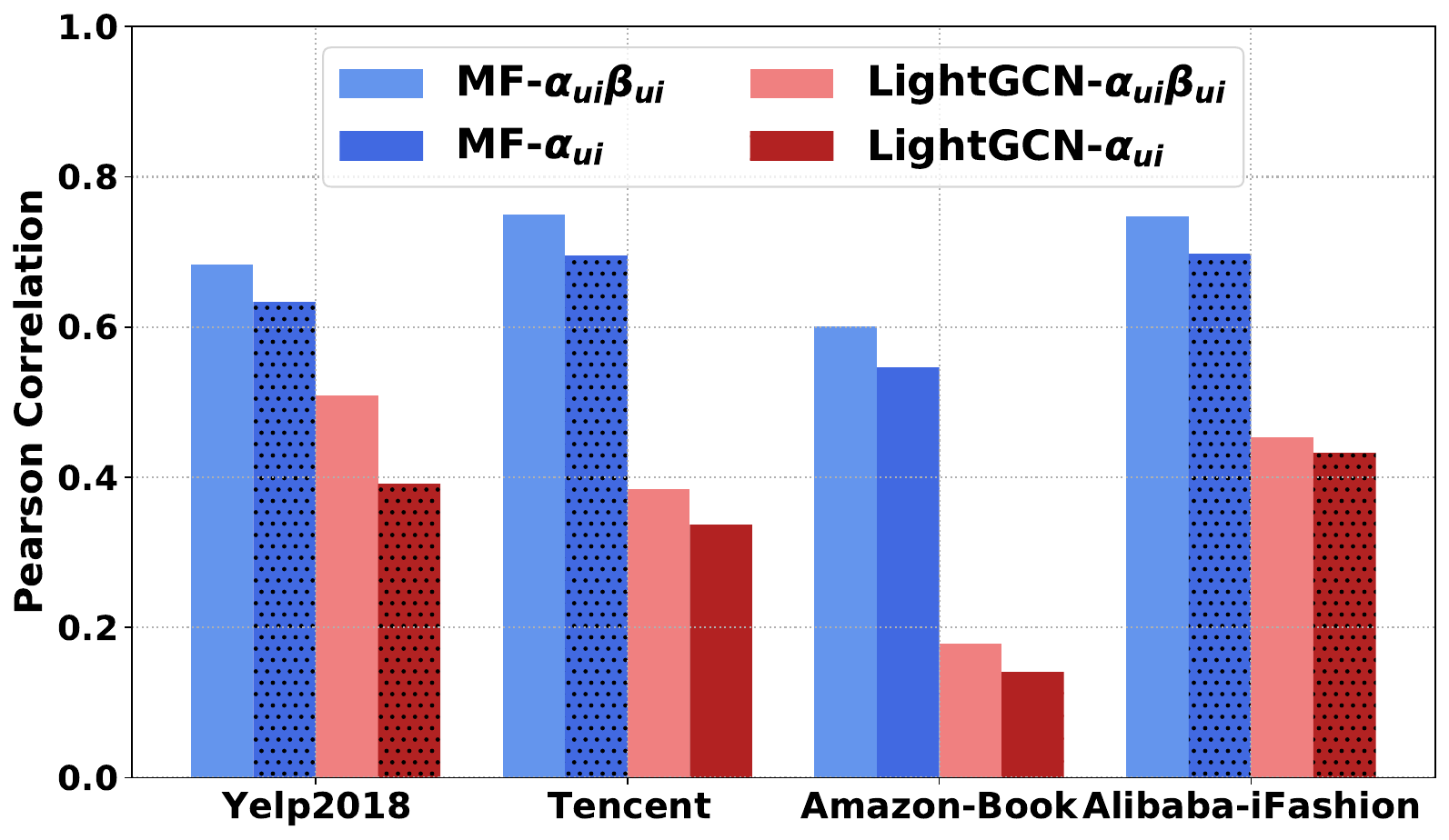}
    % \vspace{-10pt}
    % \caption{Correlation analysis of losses.}
    \caption{Correlation analysis on four datasets with both MF and LightGCN as the target models. Light color represents the correlation between the element-wise loss of dibiased models using $\alpha_{ui}$ as prediction and the shortcut model. Dark color denotes the correlation between the loss using masked prediction $\alpha_{ui}\cdot \beta_{ui}$ and the shortcut model.}
    \label{fig:cor_analysis}
    % \vspace{-20pt}
\end{figure}

To better understand the role of the shortcut model, we compare PopGo with a variant, PopGo-S that disables the shortcut model.
\begin{itemize}
    \item \textbf{Ablation Study.} Table \ref{tab:ablation1} shows that removing the shortcut model causes a huge performance drop on both ID and OOD tests. This verifies the rationality and effectiveness of the shortcut model.
    
    \item \textbf{Correlation Analysis.} With the shortcut loss in Equation \eqref{equ:shortcut-loss}, we calculate its Pearson correlation with the loss using the debiased prediction $\alpha_{ui}$ and the loss using shortcut-involved prediction $\alpha_{ui}\cdot\beta^{*}_{ui}$.
    Figure \ref{fig:cor_analysis} displays that the correlation between $\Lapl_{b}$ and $\alpha_{ui}$-loss is smaller than that between $\Lapl_{b}$ and $\alpha_{ui}\cdot\beta^{*}_{ui}$-loss.
    It verifies that $\alpha_{ui}$ holds less popularity-relevant information, thus illustrating that PopGo indeed is effective at reducing the popularity bias.
    
    \item \textbf{Visualizations of Representations.} In Figure \ref{fig:visual-embeds}, we visualize the item representations learned by MF, MF+PopGo-S, and MF+PopGo on Amazon-Book via t-SNE \cite{t-sne}. We find that: (1) The item representations of MF are chaotically distributed and difficult to distinguish, which indicates that MF focuses more on popular shortcuts, rather than preference-relevant features, which is consistent with DICE \cite{DICE};
    (2) From MF+PopGo-S to MF+PopGo, the representations present a clearer boundary, which are well scattered based on item popularity rather than chaotically distributed.
    This justifies that PopGo can learn high-quality representations for both popular and unpopular items, revealing why PopGo could achieve significant improvements in both ID and OOD settings.
    To sum up, we can confirm the impact of PopGo's shortcut model on improving the representation ability and generalization of CF models.

\end{itemize}

%% file: chapters/2_related_work.tex
\section{Related Work}
Existing debiasing strategies in CF roughly fall into five groups. 

\textbf{Inverse Propensity Score (IPS) methods} \cite{IPS,Propensity_SVM-Rank,IPS-C,IPS-CN,UBPR,Causal_inference,AutoDebias,BRD,DR} view the item popularity in the training set as the propensity score and exploit its inverse to re-weight loss of each instance. Hence, popular items are imposed lower weights, while the importance of long-tail items is boosted. However,  \cite{IPS,Propensity_SVM-Rank} suffer from the severe fluctuation of propensity score estimation due to their high variance of re-weighted loss. \cite{IPS-CN, IPS-C} further employ normalization or smoothing penalty to attain more stable output. Although IPS variants guarantee low bias, they use item frequency as a one-dimensional propensity score to capture only item-side popularity bias. This makes them fail to identify the actual interaction-wise and target model-related popularity bias. 
Recently, AutoDebias ~\cite{AutoDebias} leverages a small set of uniform data to learn debiasing parameters. 
BRD ~\cite{BRD} obtains the boundary of the true propensity score before optimizing the model with adversarial learning.
DR ~\cite{DR} proposes a doubly robust estimator to correct the deviations of errors inversely weighted with propensities. 

\textbf{Domain adaption methods} \cite{CausE,KDCRec,ESAM} utilize a small part of unbiased data as the target domain to guide the debiasing model training on biased source data. For example, CausE \cite{CausE} forces the two representations learned from unbiased and biased data similar to each other. The recently proposed KDCRec \cite{KDCRec} leverages knowledge distillation to transfer the knowledge obtained from biased data to the model of unbiased data. However, the decomposition of training data leads to two limitations - the debiased training set is often relatively small, and domain-specific information may lose. Thus, these drawbacks further result in a downgrade performance.

\textbf{Causal embedding methods} \cite{DICE,MACR,DecRS,PDA,DecCaus,CauSeR,KDCRec}, getting inspiration from the recent success of causal inference, specify the role of popularity bias in assumed causal graphs and mitigate the bias effect on the prediction.  
DICE \cite{DICE} learned disentangled interest and conformity representations by training cause-specific data. 
MACR \cite{MACR} performs multi-task learning to achieve the contribution of each cause and perform counterfactual inference to remove the popularity bias. 
CauSeR~\cite{CauSeR} uses a holistic view to capture popularity biases in data generation as well as training stages.
Using a systematic causal view of reducing popularity bias in CF, these causal embedding debiasing methods achieve state-of-the-art performance. However, despite the great success, Many of them assume the test distribution is known in advance and leverage this prior to adjusting the key hyperparameters. 

\textbf{Regularization-based methods} \cite{ESAM,ALS+Reg,Regularized_Optimization,FPC} regularize the correlations of predicted scores and item popularity, so as to control the trade-off between recommendation accuracy and coverage.
The major difference among these regularization methods is the design of penalty terms.
For example, to measure the lack of fairness, ALS+Reg \cite{ALS+Reg} introduces an intra-list binary unfairness penalty; to achieve the distribution alignment, ESAM \cite{ESAM} builds the penalty upon the correlation between high-level attributes of items; Reg \cite{FPC} decouples the item popularity with the predictive preferences;
more recently, SAM+REG \cite{Regularized_Optimization} regulates the biased correlation between user-item relevance and item popularity;
Nonetheless, most of them sacrifice overall accuracy to promote the ranking of tail items.

\textbf{Generalized methods} \cite{ bc_loss,InvCF,CD2AN,S-DRO,MIRec,COR,CausPref,InvPref} aim to learn invariant representations against popularity bias and achieve stability and generalization ability.
s-DRO ~\cite{S-DRO} adopts Distributionally Robust Optimization (DRO) framework, CD$^{2}$AN ~\cite{CD2AN} disentangles item property representations from popularity under co-training networks.
CausPref ~\cite{CausPref} utilizes a differentiable structure learner in order to learn invariant and causal user preference. 
COR~\cite{COR} formulates feature shift as an intervention and performs counterfactual inference to mitigate the effect of out-of-date interactions.
BC Loss \cite{bc_loss} incorporates popularity bias-aware margins to achieve better generalization ability.
InvCF \cite{InvCF} discovers disentangled representations that faithfully reveal the latent preference and popularity semantics.

Our proposed PopGo also shares similarities with a recently emergent category of self-supervised learning (SSL) based Collaborative Filtering (CF) methods \cite{SGL, AdvInfoNCE, RGCF, APDA}. 
SSL-based CF methods aspire to boost representation learning in recommendation systems by utilizing the principle of contrastive learning. 
These methods frequently adopt the variant of  softmax loss as their objective function.

\begin{figure}[t]
    \centering
    \includegraphics[width=\linewidth]{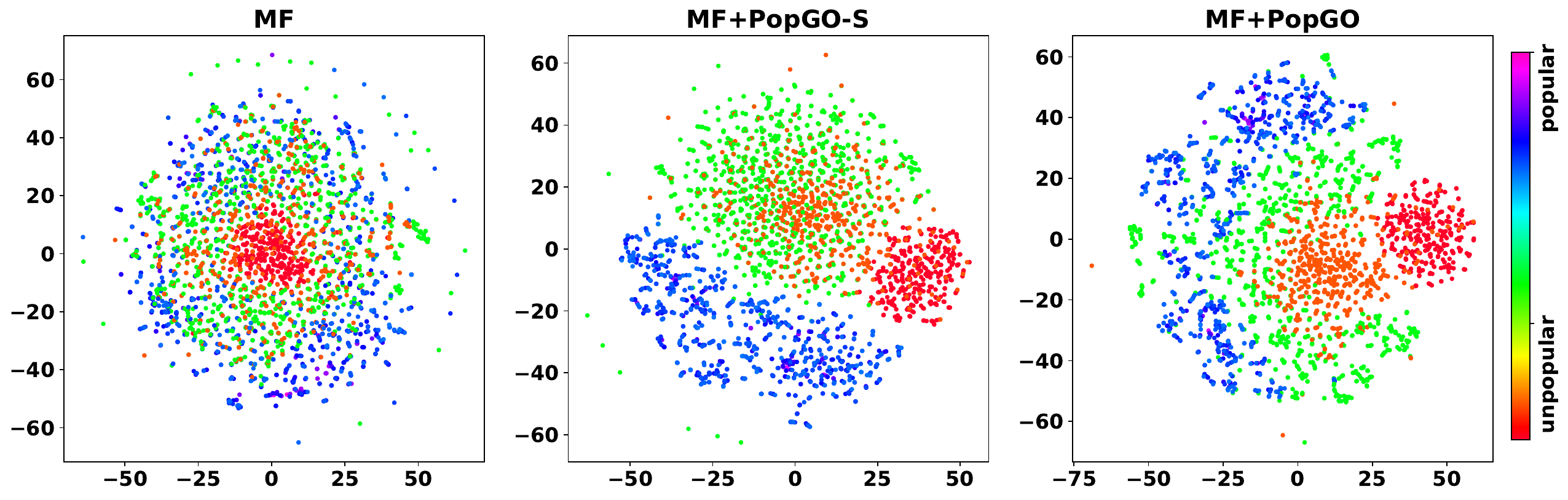}
    \vspace{-15pt}
    \caption{T-SNE \cite{t-sne} visualization of item representations.}
    \label{fig:visual-embeds}
    \vspace{-15pt}
\end{figure}

%% file: chapters/6_conclusion.tex
\section{Conclusion and Future work}
    In this paper, we proposed a simple yet effective debiasing strategy in CF, PopGo, which identifies the interaction-wise popularity shortcut degree by utilizing a shortcut model.
    Without any assumption or prior knowledge on the OOD test, PopGo achieves both ODD generalization and ID performance with high-quality debiased representations.
    Furthermore, we justify the rationality of PopGo from two theoretical perspectives: (1) from the perspective of causal inference, PopGo learns the total direct effect of user-item interactions on the prediction, while excluding the pure popularity effect;
    (2) from the perspective of information theory, PopGo intends to maximize the conditional mutual information between the interaction and prediction, conditioning on the popularity.
    Extensive experiments showcase that PopGo endows the CF models with better generalization.

PopGo primarily targets the reduction of popularity bias. An interesting avenue for future work would be to expand PopGo's capabilities to mitigate multiple biases in CF, such as exposure and selection bias. It could even tackle more challenging scenarios, like the general debiasing in recommender systems. In addition, we aim to delve deeper into the generalization of recommender models, potentially providing theoretical evidence for performance improvement in out-of-distribution settings.